\documentclass[lettersize,journal]{IEEEtran}
\usepackage{amsmath}
\usepackage{amsfonts}
\usepackage{algorithmic}
\usepackage{array}
\usepackage[caption=false,font=footnotesize]{subfig}
\usepackage{textcomp}
\usepackage{stfloats}
\usepackage{url}
\usepackage{verbatim}
\usepackage{subfig} % IEEE 推荐的子图包
\usepackage{graphicx}
\usepackage{tabularx}
\usepackage{bm}
\usepackage{booktabs} % 在文档的导言区添加这个宏包
\usepackage{siunitx} 
\usepackage{enumitem}
\usepackage{amssymb}
\usepackage{cite}
\usepackage{multirow}
\usepackage{setspace}
\usepackage{adjustbox}
\usepackage{booktabs}
\usepackage{tabu}                     % 表格插入
\usepackage{multirow}                 % 一般用以设计表格，将所行合并
\usepackage{multicol}                 % 合并多列
\usepackage{multirow}                % 合并多行
\usepackage{makecell}                 % 三线表-竖线
\usepackage{booktabs}                 % 三线表-短细横线
\newcommand{\thickhline}{%
	\noalign {\ifnum 0=`}\fi \hrule height 1pt
	\futurelet \reserved@a \@Xhline
}

\def\BibTeX{{\rm B\kern-.05em{\sc i\kern-.025em b}\kern-.08em
		T\kern-.1667em\lower.7ex\hbox{E}\kern-.125emX}}
\usepackage{balance}
\begin{document}
	
	\title{RF-LSCM: Pushing Radiance Fields to Multi-Domain Localized Statistical Channel Modeling for Cellular Network Optimization}
	% \title{How to Use the IEEEtran \LaTeX \ Templates}
	\author{Bingsheng Peng, Shutao Zhang, Xi Zheng, Ye Xue, Xinyu Qin,\\and 
		Tsung-Hui Chang, \textit{Fellow, IEEE}
		\thanks{Manuscript created June, 2025; 
			% 		Received 7 November 2023; revised 15 May 2024; accepted 3 September
			% 		2024.Dateofpublication16January2025;dateofcurrentversion6March2025.
			% 		ThisworkwassupportedinpartbytheNationalKeyResearchandDevelopment
			% 		ProgramofChinaunderGrant2022YFB2901302,inpartbytheNationalNatural
			% 		Science Foundation of China under Grant 62425104, in part by the Shanghai
			% 		Sailing Program under Grant 24YF2717400. Recommended for acceptance by
			% 		L. Bai. (Corresponding author: Weiting Zhang.)

			Bingsheng Peng, Xinyu Qin are with the School of Science and Engineering, The Chinese University of Hong Kong, Shenzhen 518172, China, and also with Shenzhen Research Institute of Big Data, Shenzhen 518172, China (e-mail: bingshengpeng@link.cuhk.edu.cn;  xinyuqin@link.cuhk.edu.cn).
			
			 Shutao Zhang and Xi Zheng are with the Networking and User Experience Laboratory, Huawei Technologies, Shenzhen 518129, China (e-mail: shutaozhang2@huawei.com; zhengxi3@huawei.com).
			
			Ye Xue is with Shenzhen Research Institute of Big Data, Shenzhen 518172, China, and also with the School of Data Science, The Chinese University of Hong Kong, Shenzhen 518172, China (e-mail: xueye@cuhk.edu.cn).
			
			Tsung-Hui Chang is with the School of Artificial intelligence, The Chinese University of Hong Kong, Shenzhen 518172, China, and also with Shenzhen Research Institute of Big Data, Shenzhen 518172, China. (e-mail:  tsunghui.chang@ieee.org).
			% 		This work was developed by the IEEE Publication Technology Department. This work is distributed under the \LaTeX \ Project Public License (LPPL) ( http://www.latex-project.org/ ) version 1.3. A copy of the LPPL, version 1.3, is included in the base \LaTeX \ documentation of all distributions of \LaTeX \ released 2003/12/01 or later. The opinions expressed here are entirely that of the author. No warranty is expressed or implied. User assumes all risk.
			
		}
		
	}
	
	% \markboth{IEEE TRANSACTIONS ON MOBILE COMPUTING}%
	% {How to Use the IEEEtran \LaTeX \ Templates}
	
	\maketitle

	%	\title{}
	%	\title[-LSCM]{RF-LSCM: Pushing Radiance Fields to  Localized Statistical Channel Modeling for Cellular Network}
	
	%\(\mathrm{RF-LSCM}\)- Tensor Radio Frequency Radiance Filed for  Wireless Communication}

\begin{abstract}
	Accurate localized wireless channel modeling is a cornerstone of cellular network optimization, enabling reliable prediction of network performance during parameter tuning. Localized statistical channel modeling (LSCM) is the state-of-the-art channel modeling framework tailored for cellular network optimization.
    However, traditional LSCM methods, which infer the channel's Angular Power Spectrum (APS) from Reference Signal Received Power (RSRP) measurements, suffer from critical limitations: they are typically confined to single-cell, single-grid  and single-carrier frequency analysis and fail to capture complex cross-domain interactions. To overcome these challenges, we propose RF-LSCM, a novel framework that models the channel APS by jointly representing large-scale signal attenuation and multipath components within a radiance field. RF-LSCM introduces a multi-domain  LSCM formulation with a physics-informed frequency-dependent Attenuation Model (FDAM) to facilitate the cross frequency generalization as well as a point-cloud-aided environment enhanced method to enable multi-cell and multi-grid channel modeling. Furthermore, to address the computational inefficiency of typical neural radiance fields, RF-LSCM leverages a low-rank tensor representation, complemented by a novel Hierarchical Tensor Angular Modeling (HiTAM) algorithm. This efficient design significantly reduces GPU memory requirements and training time while preserving fine-grained accuracy. Extensive experiments on real-world multi-cell datasets demonstrate that RF-LSCM significantly outperforms state-of-the-art methods, achieving up to a 30\% reduction in mean absolute error (MAE) for coverage prediction and a 22\% MAE improvement by effectively fusing multi-frequency data.
\end{abstract}

\maketitle

\begin{figure*}[t]
	\centering
	\includegraphics[width=\linewidth]{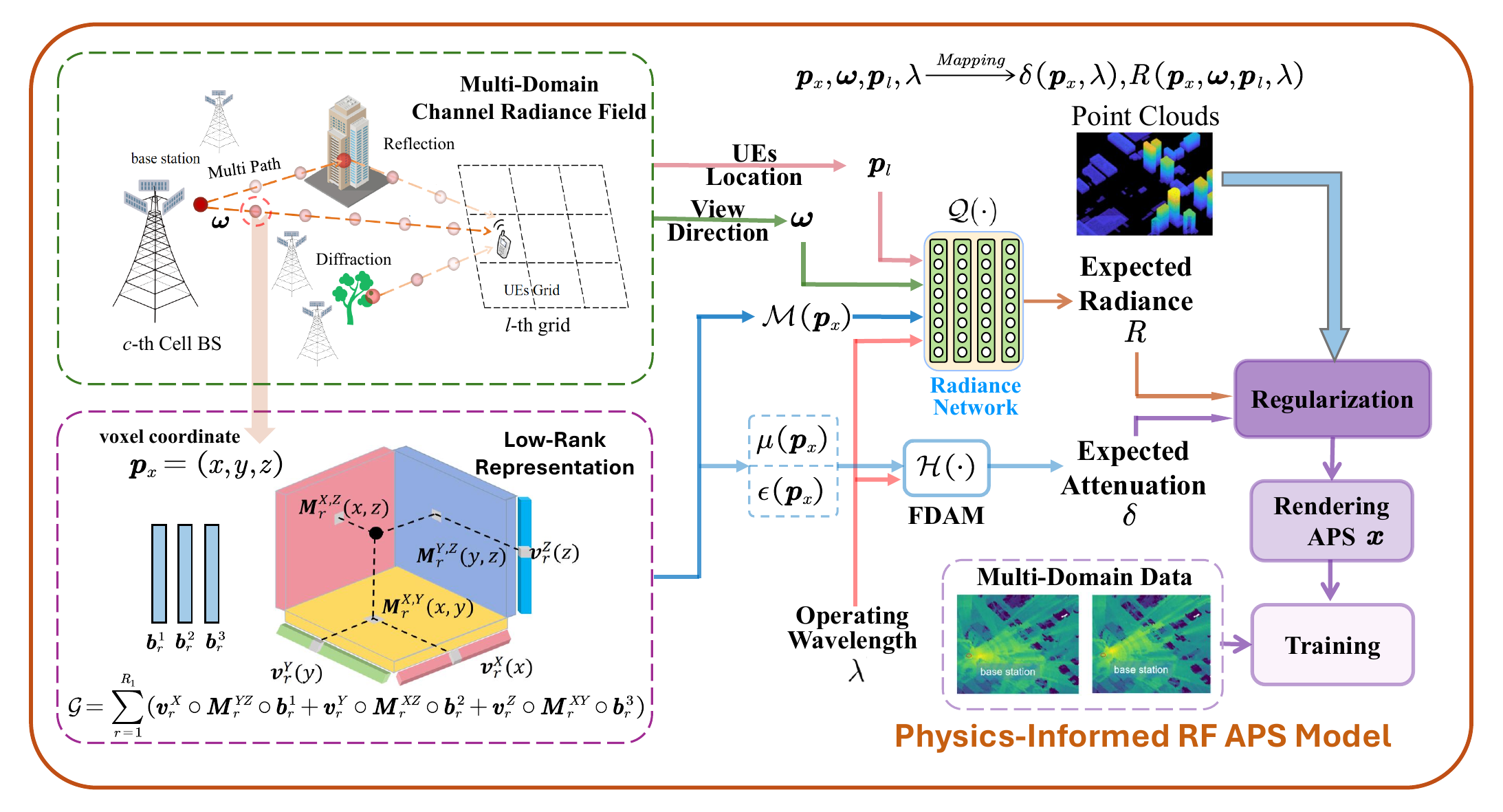}
	\caption{\textbf{System Workflow of RF-LSCM.} The framework models the cellular environment as a 3D voxel grid represented by a low-rank tensor and integrates a novel  FDAM  within a radiance field while using point clouds to regularize the radiance field parameters. This allows the system to render the channel APS, enabling accurate prediction of base station coverage performance for diverse antenna configurations.}
	\label{fig:tensor_decompsition}
	%	\Description{A visual representation of the challenges in modeling wireless signal propagation due to environmental factors.}
\end{figure*}

\section{Introduction \label{sec:Itroduction}}

The rapid advancement of the fifth-generation (5G) mobile communication technology and the ongoing exploration of the sixth-generation (6G) systems have made cellular network optimization  a central focus of research \cite{10430216}. In 5G networks, the optimization of wireless cellular networks is increasingly challenging due to the scaling-up of the number of parameters to be optimized, including those associated with antennas, handover processes, beamforming techniques, and carrier configurations \cite{10155734}. Traditional network optimization methods, which depend on engineering expertise and multiple rounds of drive tests \cite{Cao2018DesignAV,6353680}, are often labor-intensive and unable to meet the performance requirement of a large-scale cellular network. Recently, 
the concept of digital twins  \cite{9374645} has motivated the use of data-driven approaches to perform offline network optimization, avoiding risks caused by online parameter tuning. This approach allows for the modeling and simulation of wireless networks without the need of multi-round manual intervention, thereby enhancing efficiency and reducing cost in the network optimization process.

Channel modeling is one of the most fundamental components for network optimization since an accurate channel model provides reliable predictions of network performance during the parameter optimization process. 
Existing 5G channel models, including GBSMs \cite{9318511,8424015} and ray-tracing based models \cite{8438326,8630961,9504428}, are inadequate for the network optimization task. 
Specifially, GBSMs are designed for typical communication scenarios and cannot provide localized modeling for targeted scenarios, while ray tracing cannot perform well without accurate map information \cite{9373011,7498102,8046051}. Simplified empirical models such as COST231-Hata \cite{1543252} are too simple to capture the complexity and randomness of 5G networks. These limitations result in unsatisfactory optimization configurations that underperform in real-world scenarios.

To address these limitations, data and physic based localized statistical channel modeling (LSCM) \cite{10299600} has been proposed as a tailored solution for cellular network optimization. LSCM characterizes the multi-path structure of localized propagation environments by estimating the fine-grained angular power spectrum (APS) of wireless channels based on multi-beam Reference Signal Received Power (RSRP) measurements. Notably, LSCM has been successfully implemented in the SRCON platform  \cite{huawei_solving_network_mission_impossible,huawei_greening_network_with_ai}, a real-world telecom optimization tool used for 5G network. SRCON integrates statistical mechanisms and builds large-scale simulation models of real-world networks, including multi-path channels, user distribution, and base station responses. This approach has enabled significant achievements in 5G network optimization, such as leading the European P3 assessments in Spain, Germany, the Netherlands, and Switzerland \cite{sribd_srcon}.
% \begin{figure}
% 	\centering
% 	\includegraphics[width=0.9\linewidth]{plot/global_channel2}
% 	\caption{The global environmental impact on channel propagation. (a) shows that different grids may exhibit similar channel characteristics, with angles differences less than a small value \(\Delta \omega\), due to environmental reflections. (b) demonstrates that channels from different cells share the same characteristics when radio waves traverse the same space and intersect.}
% 	\label{fig:globalchannel}
% \end{figure}
\begin{figure}[!t]
	\centering
	\includegraphics[width=0.9\linewidth]{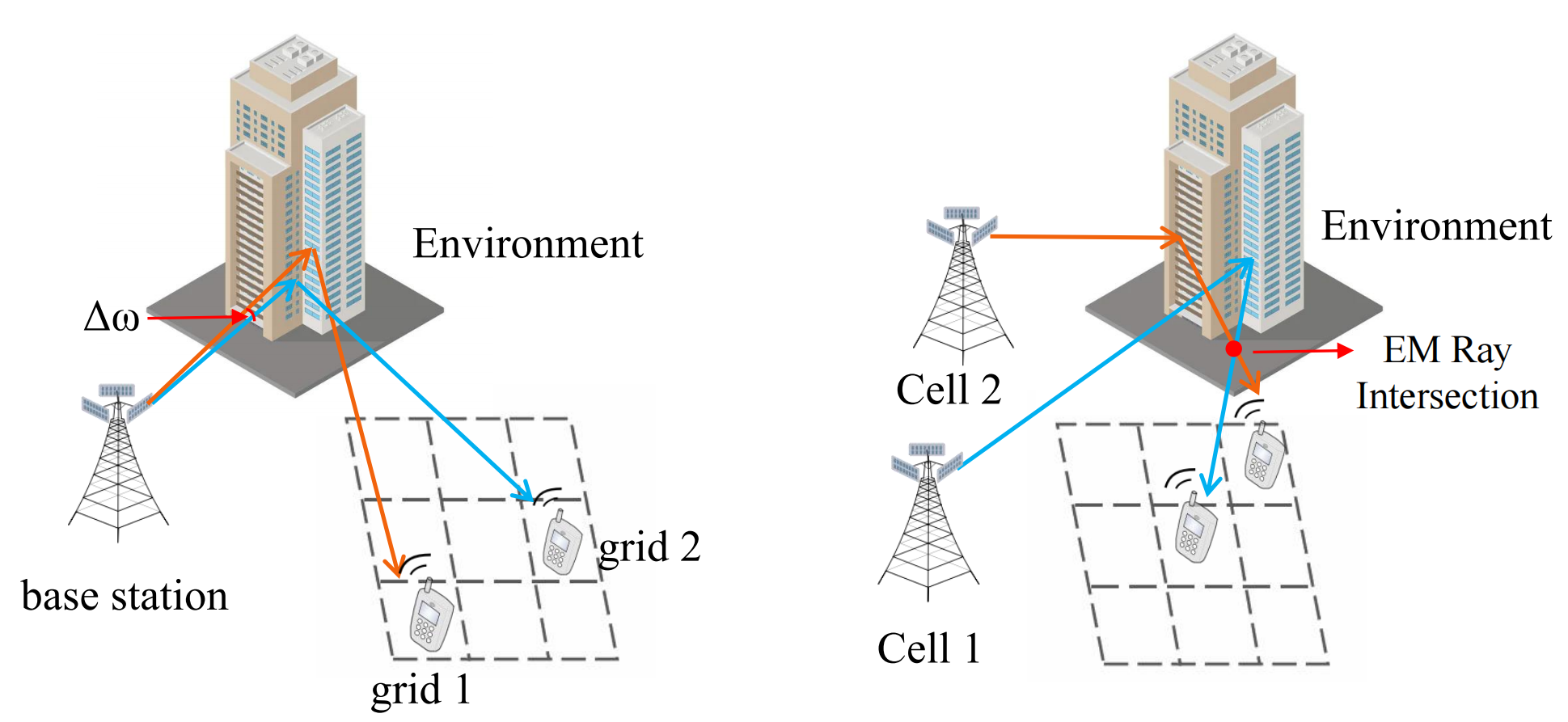}
	
	% --- 修改后的 Caption (无 a/b 标签) ---
	\caption{The global environmental impact on channel propagation. 
	On the left, the illustration shows that different grids may exhibit similar channel characteristics, with angle differences less than a small value \(\Delta \omega\), due to environmental reflections. 
	On the right, it demonstrates that channels from different cells share the same characteristics when radio waves traverse the same space and intersect.}
	\label{fig:globalchannel}
\end{figure}
However, the original LSCM approach primarily focuses on modeling the APS between a base station (BS) and a single geometric grid, only using RSRP data from that grid while ignoring the geometric relationships across different grids. 
Although subsequent works \cite{10622313,9940390} have attempted to extend LSCM to multi-grid scenarios, these methods still rely on oversimplified statistical priors  between adjacent grids, thus fail to model the intricate interactions between radio wave propagation and the physical environment, as illustrated in Fig.~\ref{fig:globalchannel}. Compounding by this issue, their applicability remains strictly confined to single-cell and single-carrier frequency scenarios. This critical limitation hinders their ability to leverage the intrinsic dependencies within \textbf{multi-domain} data (multi-carrier frequency, multi-cell, and multi-grid data), which in turn restricts their effectiveness for large-scale network optimization tasks. Consequently, there is a clear and growing necessity for an advanced localized channel modeling approach, one that can faithfully characterize the influence of the environment on radio wave propagation and truly address the limitations of existing methods.

Recent developments in deep learning have markedly enhanced the simulation of radio-wave propagation, offering great promise for more accurate channel modeling. A recent state-of-the-art method, named neural radio-frequency radiance fields (NeRF$^2$)  \cite{Zhao_2023}, employs a neural radiance field (RF) to model the radio signal of indoor scenarios by tracing rays from all possible directions. However, NeRF$^2$ primarily focuses on radio map construction, not on the fine-grained APS modeling required for LSCM.
APS modeling from RSRP is intrinsically an ill-imposed estimation problem and is more challenging to solve than the tasks considered in \cite{Zhao_2023}. Moreover, while this neural radiance field method has shown strong potential, its direct application to outdoor cellular networks poses substantial challenges. Outdoor environments involve significantly larger service areas, multiple cells, and more complex propagation characteristics compared to indoor settings. The training process for its deep multi-layer perceptron (MLP) architecture requires extensive computational resources\cite{mueller2022instant}, a necessity driven by the processing of large-scale outdoor datasets, which in turn demands substantial GPU memory and time. These challenges are further exacerbated when modeling fine-grained APS for LSCM in outdoor environments, as this task entails capturing detailed angular variations over broad spatial areas for wireless network optimization while accounting for increased multi-path complexity. The combination of larger spatial scales, higher angular resolution requirements, and the intricate physical structure of outdoor environments renders this neural radiance field method impractical for cellular networks without significant adaptations. Consequently, a novel radiance field-based approach that effectively captures the interactions between radio waves and the physical environment is essential to overcome these limitations and enable LSCM in cellular networks.

% \subsection{Motivations and Contributions}
To address the aforementioned issues, we propose RF-LSCM (\underline{R}adiance \underline{F}ield-based \underline{L}ocalized \underline{S}tatistical \underline{C}hannel \underline{M}odeling for {Ce}llular Networks),   a novel framework designed for efficient and accurate channel modeling in complex multi-cell, multi-carrier frequency environments. 
 Fig.~\ref{fig:tensor_decompsition} displays the system workflow of the proposed RF-LSCM.
At its core, RF-LSCM introduces a new paradigm by directly representing the channel APS within a radiance field. The framework formulates this as a multi-domain APS recovery problem, based on the physical assumption that wave propagation can be decomposed into an expected radiance and an expected attenuation. To model the latter, we introduce our Frequency-dependent  Attenuation Modeling (FDAM), which accurately captures frequency-dependent attenuation. Furthermore, to ensure the model's physical fidelity, we leverage point cloud data as a geometric prior to regularize both the radiance and attenuation fields against the physical environment. This holistic design enables RF-LSCM to effectively capture fine-grained spatial and angular dependencies by efficiently fusing multi-frequency, multi-cell, and point cloud data, thereby significantly improving coverage prediction. 

Crucially, to overcome the computational bottlenecks of traditional methods, RF-LSCM's architecture achieves its efficiency by synergistically combining two key architectural strategies. First, we employ a low-rank Vector-Matrix (VM) tensor decomposition technique \cite{10.1007/978-3-031-19824-3_20}, replacing the expensive MLP with a small number of highly efficient tensor operations. Second, we introduce Hierarchical Tensor Angular Modeling (HiTAM), 
a novel method designed to not only mitigage the complexity arising from high angular granularity but also enhance the model learning performance. HiTAM begins with a coarse-grained APS followed by a refined angular granularity, which effectively reduces the number of angular powers to compute and simultaneously alleviates the problem’s ill-posedness for APS prediction.
Collectively, these strategies empower RF-LSCM to maintain high-fidelity APS modeling while achieving the computational efficiency required for practical, large-scale deployment. 

The main contributions of this paper are summarized as follows:
\begin{itemize}
	\item \textbf{A Physics-Informed, Multi-Domain Channel Model.} We propose RF-LSCM, a novel framework that directly models the  multi-frequency, multi-cell and multi-grid channel APS within the radiance field. By incorporating our physics-informed FDAM, RF-LSCM can explicitly model the physical impact of the environment on wave propagation, rather than merely fitting to received signal strength. This approach creates a physically interpretable and robust foundation for network optimization in complex outdoor environments.
	
	\item \textbf{An Efficient Tensor-Based Radiance Field Architecture.} To circumvent the prohibitive computational cost of traditional MLP-based methods, we introduce a highly efficient, tensorized RF representation. Our design synergistically combines low-rank tensor decomposition with a novel HiTAM algorithm. This two-pronged approach drastically reduces GPU memory overhead and training time while preserving the fine-grained accuracy essential for detailed channel modeling.
	
	\item \textbf{State-of-the-Art Performance on Real-World Datasets.} Extensive experiments conducted on large-scale, real-world multi-cell, multi-frequency datasets confirm that RF-LSCM achieves state-of-the-art performance. Our method significantly outperforms existing approaches, yielding up to a 30\% reduction in RSRP prediction MAE. Furthermore, the model's multi-domain capability is underscored by a 21.8\% MAE reduction achieved through the fusion of multi-frequency data, highlighting the effectiveness of this modeling strategy.
\end{itemize}

This paper is organized as follows. Section \ref{sec:problem statement} introduces the system model. Section \ref{sec:RF_model} details the proposed RF-LSCM. In Section \ref{sec:complex_reduciton}, we describe the complexity reduction methods for APS modeling. Section \ref{sec:evaluation} presents real-world test results that demonstrate our model's efficacy. Subsequently, Section \ref{sec:ablation_study} shows an ablation study analyzing the impact of key components. Finally, Section \ref{sec:concluding} concludes the paper with a summary of our work. Key parameter notations are listed in Table~\ref{table:parameters}.
\begin{table}[t]
	\small
	\caption{Summary of Key Notations.}
	\centering
	
	% Using booktabs for a professional look.
	% No vertical lines, and well-spaced horizontal rules.
	\begin{tabularx}{\columnwidth}{@{} l X @{}} % l: left-align, X: auto-wrap, @{}: removes side padding
		\toprule
		\textbf{Notation} & \textbf{Definition} \\
		\midrule
		% High-Level & APS Concepts
		\(M\) & Number of measured RSRP beams. \\
		\(\bm{r} \in \mathbb{R}^M\) & Vector of measured RSRP values. \\
		\(N\) & Number of angular divisions for AoD. \\
		\(\bm{x} \in \mathbb{R}^N\) & High-resolution Angle of APS. \\
		\(\bm{\tilde{x}}\) & Reconstructed (approximated) APS. \\
		\midrule
		% Scene & Indices
		\(L, C\) & Total number of geometric grids and serving cells. \\
		\(\bm{p}_{BS}\) & Position of the base station. \\
		\(\bm{p}_x\) & Position of a voxel. \\
		\(\bm{\omega}\) & Direction of EM ray. \\
		\(n, l, c\) & Indices for angle, grid, and cell. \\
		\midrule
		% Physical & Field Parameters
		\(\mu, \epsilon\) & Complex relative permeability and permittivity. \\
		\(\bm{\theta}\) & EM parameter vector per voxel, \(\bm{\theta} \triangleq (\mu, \epsilon)\). \\
		\(\delta\) & Expected attenuation for a voxel. \\
		\(R\) & Expected radiance for a voxel. \\
		\midrule
		% Field Representation & ML
		\(\bm{\mathcal{G}}_{\delta}, \bm{\mathcal{G}}_{R}\) & 4D tensors for the attenuation and radiance fields. \\
		\(\bm{v}, \bm{M}\) & Vector and Matrix components from VM decomposition. \\
		$\mathcal{Q}$  & Small MLP network for expected radiance prediction. \\
		\bottomrule
	\end{tabularx}
	\label{table:parameters}
	\vspace{-4pt}
\end{table}
\section{System Model and Problem Formulation}
\label{sec:problem statement}
%
%======================================================================================
\subsection{Fundamentals of LSCM }
\label{sec:LSCM}
%======================================================================================
%
%In modern wireless networks, a channel model that accurately captures the environment's specific multi-path topography is essential for various optimization tasks \cite{10155734}. The  LSCM framework \cite{10299600} was introduced to estimate the channel APS  from beam-wise  RSRP measurements.

The vanilla LSCM \cite{10299600} considers a single-cell scenario where the BS is equipped with a uniform rectangular antenna array of \( N_T = N_x \times N_y \) elements, and each User Equipment (UE) is equipped with a single antenna. The coverage area is divided into grids, with UEs assumed to be located within the grids.
The BS periodically transmits   synchronization Signal Blocks (SSB) and Channel State Information (CSI) reference beam signals to the UEs. {Suppose that the BS transmits \( M \) directional beams, and let the precoding matrix for the \( m \)-th beam be \( \mathbf{W}^{(m)} \in \mathbb{C}^{N_x \times N_y} \). Let \(h_{x,y}\) be the complex channel coefficient from the antenna at index \( (x, y) \) to a UE, where \( x \in \{ 0, \ldots, N_x-1 \} \) and \( y \in \{ 0, \ldots, N_y-1 \} \).} The expected RSRP at the grid due to this beam is given by
\begin{equation}
	\text{RSRP}_m \triangleq \mathbb{E} \left[ P_t \left| \sum_{x=0}^{N_x-1} \sum_{y=0}^{N_y-1} h_{x,y} W_{x,y}^{(m)} \right|^2 \right],  
\end{equation}
where \( \mathbb{E}[\cdot] \) is the expectation over the channel realizations, \( W_{x, y}^{(m)} \) is the \( (x, y) \)-th entry of \(\mathbf{W}^{(m)}\), and \( P_t \) is the BS transmit power. The collection of all \(M\) RSRP measurements forms the vector
\begin{equation}
	\bm{r} = \left[ \text{RSRP}_1, \text{RSRP}_2, \ldots, \text{RSRP}_M \right]^T.
\end{equation}

%In wireless communications, the channel coefficient \(h_{x,y}\) is usually modeled to have a multi-path structure, each characterized by an angle of angle of departure (AoD) and a path power. To model this,  the spatial angular domain is uniformly discretized into \( N \) directions, with \(N\) typicall being significantly larger than M to achieve a high angular resolution to capture the high-grain channel information . As demonstrated in \cite{10299600}, the RSRP measurements \(\bm{r}\) and the channel APS \(\bm{x} \in \mathbb{R}^N\) are linked by a linear relationship:
In wireless communications, the channel \(h_{x,y}\) is physically composed of multiple propagation paths, each characterized by a distinct Angle of Departure (AoD) and path power. To capture this multi-path structure, the 2-dimensional  angular domain is discretized into \(N\) uniformly spaced bins, each specified by an azimuth and elevation angle, and the number of bins \(N\) is significantly larger than the number of RSRP measurements  \(M\) (\(N \gg M\)). 
Let \( \bm x \in \mathbb{R}^N\) denote the channel path power across the angular bins. It has been established in \cite{10299600} that the RSRP measurements \(\bm{r}\) and the APS \(\bm{x}\) are linked by the linear relationship
\begin{equation}
	\bm{r} = \mathbf{A}{\bm{x}} \label{eq:yax},
\end{equation}
where \( \mathbf{A} \in \mathbb{R}^{M \times N} \) is the sensing matrix determined by the antenna configuration and beamforming waveforms. Due to the limited number of significant scatterers in the environment, the APS vector \(\bm{x}\) is typically sparse. Recovering \(\bm{x}\) from \(\bm{r}\) is thus framed as a sparse recovery problem\cite{10299600}
\begin{equation}
	\label{eq:muticell}
	\begin{aligned}
		\underset{\bm{x}}{\min} & \quad \left\| \bm{r} - \mathbf{A} \bm{x} \right\|_2^2 \\
		\text{subject to} & \quad \left\| \bm{x} \right\|_0 < K, \\
		& \quad x_n \geq 0, \quad \text{for } n = 1, \dots, N,
	\end{aligned}
\end{equation}

where \( \|\cdot\|_0\) denotes \(\ell_0\) norm. 
This formulation seeks to obtain an APS solution for the grid that minimizes the data fidelity term while enforcing sparsity (\(\left\| \bm{x} \right\|_0 < K\)) and non-negativity (\(x_n \geq 0\)) constraints of channel power spectrum.  
In \cite{10299600}, several orthogonal matching pursue (OMP) type algorithms are proposed to handle the challenging problem (\ref{eq:muticell}).

\subsection{Problem Formulation }
\label{sec:multi-domain-modeling} 
Treating channel model as an independent inverse problem for each grid, as formulated in problem (\ref{eq:muticell}), is strictly limited. This is because it considers the RSRP measurements from one BS only and estimating the APS of a grid at a time. Since a UE can receive signals from multiple cells simultaneously and different cells can operate at different carrier frequencies, such a single-domain modeling in (\ref{eq:muticell}) thereby fails to exploit the strong underlying correlations that exist across cells, grids and carrier frequencies.
 % 
% Paragraph 2: The Physical Roots of Correlation
In particular, these correlations, which conventional LSCM overlooks, manifest in several ways:
\begin{itemize}
	\item \textbf{Spatial Correlation}: The propagation channels from two different BSs to the same grid point are not independent; As shown in Fig.~\ref{fig:globalchannel}, they are shaped by the same local scatterers and blockers, just viewed from different angles. Similarly, the APS for nearby grid points from the same BS should be highly similar, and likely to change smoothly with location.
	\item \textbf{Frequency Correlation}: The channel for the same cell-grid link but at different carrier frequencies are also deeply related. Physical phenomena such as diffraction, scattering, and material penetration are inherently frequency-dependent, leading to predictable structural similarities in the APS across bands.
\end{itemize}

% Paragraph 3: The Solution Concept - A Unified, Shared Representation
Therefore, the central challenge is not merely to solve an array of independent inverse problems, but to develop a unified, physically-informed model. The core idea is to learn a single, shared representation of the environment. By doing so, the model can intrinsically capture these complex cross-domain dependencies and generate a consistent  APS \(\bm{   x}\)  for any cell \(c\), grid \(l\), and carrier wavelength \(\lambda\),  i.e.,
\begin{equation}
	\bm{r}_{c,l}^\lambda = \mathbf{A}_{c}^\lambda \bm{  x}(c, l, \lambda ) \label{eq:multi-domain formulation}.
\end{equation}
%
%\subsection{Neural representation of Environment}
% Paragraph 4: The New Hurdles - Conceptual and Computational
The vision of a unified environmental model is powerfully inspired by recent advances in neural rendering methods, which excel at learning shared 3D scene representations \cite{Zhao_2023}. However, adapting this paradigm to large-scale channel modeling introduces two fundamental challenges. 

Physics-consistency challenge: Adapting a framework originally built for 3D vision reconstruction to accurately represent the unique physics of radio wave propagation—including multi-domain APS in radiance fields across various cells and frequencies—requires a substantial conceptual overhaul, as the connection is not straightforward.

Computational challenge: The inverse problem outlined in  (\ref{eq:muticell}) is highly ill-posed (with \(N\) far exceeding \(M\)), complicating efficient optimization, while the demands of modeling high-dimensional APS using standard neural field approaches lead to excessive memory and processing needs, rendering them impractical for expansive outdoor wireless environments.

% Paragraph 5: The Roadmap - Our Proposed Solution
To address these challenges---the need for a physically-unified model and the demand for computational efficiency---this paper proposes RF-LSCM. In the following sections, we will first detail this physics-informed APS model. At its core, RF-LSCM introduces a novel parameterization of the APS as a function of the environment's fundamental electromagnetic (EM) properties. This parameterization is the key to effectively fusing multi-domain information. Then, in Section~\ref{sec:complex_reduciton}, we present our techniques to 
improve the model optimization performance while mitigating the associated computational complexity. Specifically, we propose a hierarchical estimation algorithm that begins with a coarse-grained APS and subsequently refines it to a fine granularity. This approach effectively reduces the problem's ill-posedness and improves model speed by decreasing the number of angular powers to compute. Additionally, we leverage a tensor-based methodology to enhance both model performance and efficiency.

\section{Physics-Informed Multi-domain APS Model \label{sec:RF_model}}
In this section, we detail the architecture of RF-LSCM, our framework for modeling multi-domain APS within a radiance field representation. The framework operates on the physical principle that wave propagation can be decomposed into two fundamental components: an expected radiance and an expected attenuation. To enable robust cross-frequency generalization, we introduce our FDAM, which accurately captures how attenuation varies with signal wavelength. Furthermore, to anchor the model in physical reality, we leverage point cloud data as a strong geometric prior to regularize both the radiance and attenuation fields. Finally, we explain how the model is optimized to recover a  multi-domain APS representation by learning the environment's fundamental EM parameters from diverse multi-cell and multi-frequency data.
\subsection{Radiance Field APS Model}
%\textbf{Modeling the Environment as a Radiance Field.}
%\subsubsection{Multi-domain APS Modeling}
%
Based on the principle of channel reciprocity \cite{10.1145/3447993.3483275}, we model the BS as a receiver and UEs as transmitters, denoting the Angle of Arrival (AoA) by \(\bm{\omega}\). Inspired by recent advances in the use of neural rendering for RF model \cite{Zhao_2023}, we represent the propagation environment as a field of voxels. Since our goal is to model the statistical APS, we conceptualize each voxel as a transmitter and blocker, with each voxel being characterized by two fundamental, complex-valued {quantifies} that are determined solely by the propagation environment: \textbf{expected attenuation} (\(\delta\)), which quantifies the average signal power reduction due to obstructions, and \textbf{expected radiance} (\(R\)), which represents the voxel's average ability to scatter or emit signals. 

{The channel power at wavelength \(\lambda \)  from a BS cell to a grid along a direction \(\bm \omega\) is formulated by modeling the aggregate contributions of all voxels along the corresponding path. This path is modeled as an EM ray, uniformly discretized into \(Q\) voxels. The contribution of each voxel \(q\) is its expected radiance \( {R}_{ q }^{\lambda }\), attenuated by the cumulative transmittance of the voxel itself and all preceding ones. 
The expected signal \(\tilde s^\lambda(\bm \omega_n)\) and the expected channel power \(\tilde x^\lambda(\bm \omega_n) \) along the AoA direction  \(\bm \omega_n\) can be modeled as
\begin{equation}
	\tilde s^\lambda(\bm \omega_n)  =   \sum_{q=0}^Q \left( \prod_{\tilde{q}=0}^q  {\delta}_{   \tilde{q}}^{\lambda } \right)  {R}_{  q}^{\lambda }   , \label{eq:s_render}
\end{equation}
\begin{equation}
	\tilde x^\lambda(\bm \omega_n)  = \left| \tilde s^\lambda(\bm \omega_n) \right|^2, \label{eq:x_render}
\end{equation}
%Sampling this across all \(N\) discrete angles yields the modeled APS vector \(\bm{\tilde{x}}_{c,l}^{\lambda_c}\):
%\begin{equation}
%	\bm{\tilde{x}}_{l}^{\lambda} = [\tilde{x}_{l,\lambda}^{1}, \ldots, \tilde{x}^{N}_{\lambda}]^T. \label{eq:x_vec}
%\end{equation}
where \( \prod_{\tilde{q}=0}^q \delta_{\tilde q}^\lambda\) is the cumulative product of expected transmittance factors along the propagation path. Sampling across all \(N\) discrete AoA directions yields the complete APS prediction vector
\(\bm {\tilde x }= [ \tilde x^\lambda(\bm \omega_1), \ldots, \tilde x^\lambda(\bm \omega_n), \ldots,\tilde x^\lambda(\bm \omega_N)]\).
Then, by (\ref{eq:muticell}), the RSRP vector \(\bm r\) can be estimated by \(\bf A \bm \tilde x\). It is important to note that this is distinct from  NeRF$^2$  \cite{Zhao_2023} that uses (\ref{eq:x_render}) to predict the signal value directly via \( \sum_{  n=1}^N \tilde s^\lambda(\bm \omega_n)\). 
%It is important to clarify that our model predicts the  {channel power}, which is the expected squared magnitude of the channel gain. This value is a normalized, dimensionless quantity that characterizes the intrinsic effects of the propagation environment itself. This is distinct from predicting the absolute received  {signal power} in NeRF$^2$ \cite{Zhao_2023}, which would additionally depend on system-specific parameters such as the transmitter's power and antenna gains.
%
%
%that \(\bm{\tilde{x}}\) is an approximation of the true APS. Our model calculates the squared magnitude of the coherent sum of path contributions (i.e., the expected channel gain, squared), whereas the true APS is the non-coherent sum (i.e., the expected value of the squared channel gain).}
%\begin{equation}
%	\bar{x} = \left| \sum_{q=0}^Q \left( \prod_{\tilde{q}=0}^q  {\delta}_{ \tilde{q}}  \right)  {R}_{ q }  \right|^2. \label{eq:x_start}
%\end{equation}
%
\subsection{Frequency-dependent  Attenuation Modeling (\textbf{FDAM})}
The primary objective of our  FDAM is to build a physically-informed representation of the expected attenuation \(\delta\), that enables robust cross-frequency generalization. Rather than modeling \(\delta\) as a direct mapping from voxel coordinates \cite{Zhao_2023}---we posit that \(\delta\) is a function of more fundamental EM properties of the medium, which are inherently dependent on the wavelength \(\lambda\). 

\begin{figure}
	\centering
	\includegraphics[width=1\linewidth]{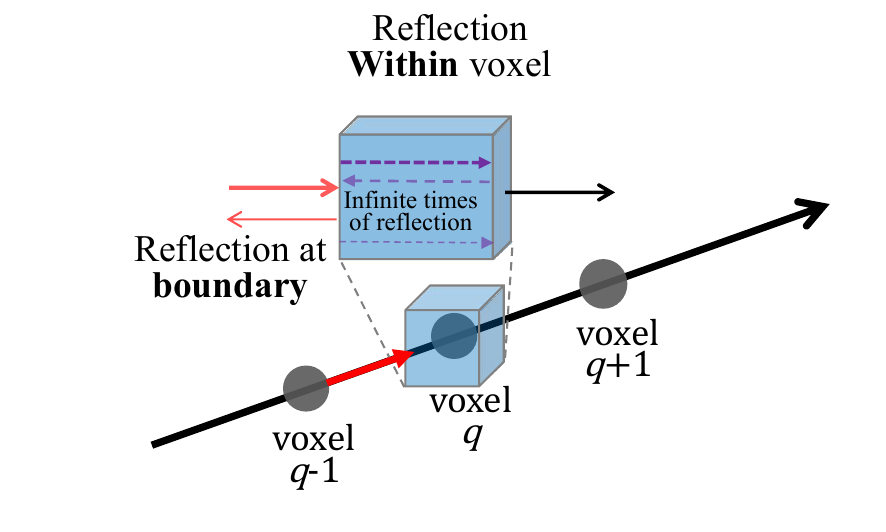}
	\caption{Illustration of FDAM. EM rays exhibit  attenuation both in the boundaries and within the interior regions. }
	\label{fig:mvam}
\end{figure}
Consider an EM wave propagating along a ray that is discretized into a sequence of \(Q\) contiguous voxels. As illustrated in Fig.~\ref{fig:mvam}, the wave's interaction with this voxelized medium is governed by two primary phenomena.
%
%We first establish a physical-informed model for expected attenuation within a single EM ray. Instead of  modeling \(\delta\) by the directly using a mapping function from the voxel coordinate, we here consider the \(\delta\) is modeled as a function of the 更本质的 EM parameters related to wavelenght \(\lambda\). Consider an EM wave propagating along a ray that is discretized into a sequence of \(Q\) contiguous voxels. The wave's interaction with this medium is governed by two phenomena, as illustrated in Figure~\ref{fig:mvam}.
%
First, at the interface between adjacent voxels (e.g., from voxel \(q-1\) to voxel \(q\)), impedance discontinuities cause partial reflection, quantified by the Fresnel reflection coefficient \( \Gamma_{q} \) \cite{balanis2012advanced}
\begin{align}
	\Gamma_{q} = \frac {Z_{q} - Z_{q-1} }{Z_{q} + Z_{q-1}}, \label{eq:Gamma_q_vivid}
\end{align}
where \( Z_q = \sqrt{\mu_q / \epsilon_q} \) is the normalized characteristic impedance of voxel \(q\), and \(\mu_q\) and \(\epsilon_q\) are its complex relative permeability and permittivity, respectively.
Second, as the wave traverses a single voxel \(q\), its phase and amplitude are altered. This is described by the internal propagation factor \( T_{q}^{\lambda} \) \cite{balanis2012advanced}, which depends on the voxel's   \( \epsilon_{q} \),   \( \mu_{q} \), and the wavelength \( \lambda \)
\begin{align}
	T_{q}^{\lambda} = \exp\left(-j\frac{2\pi}{\lambda}\sqrt{\epsilon_{q} \mu_{q}} d\right). \label{eq:T_vivid}
\end{align}
The total attenuation of voxel \(q\), \(\delta_q \), is not merely internal propagation, but instead it holistically accounts for the initial transmission into the voxel followed by a cascade of internal reflections between its boundaries, analogous to a Fabry-Pérot etalon \cite{BornWolf:1999}. By summing this infinite geometric series, we obtain the FDAM:
\begin{align}
	{\delta}_q^\lambda 	&= \underbrace{(1 - \Gamma_{q})}_{\substack{\text{pass through} \\ \text{boundary}}}
	\underbrace{T_{q}^{\lambda}}_{\substack{\text{pass through} \\ \text{voxel}}} 
	\underbrace{\sum_{k=0}^{\infty} \left(-\Gamma_{q}\Gamma_{{q+1}}(T^{\lambda }_{q})^2\right)^k}_{\text{infinite times of reflections}}    \label{eq:all_attenuation}  \\
	&= \frac{(1 - \Gamma_{q}) T^{\lambda }_{q}}{1 + \Gamma_{q}\Gamma_{q+1}(T^{\lambda }_{q})^2} \triangleq \mathcal{H}(\bm\theta_{q-1}, \bm\theta_{q}, \bm\theta_{q+1}, \lambda), \label{eq:mvam_vivid}
\end{align}
where the three terms in the right hand side of (\ref{eq:all_attenuation}) are respectively due to that the signal passes through the boundary between voxel \(q-1\) and voxel \(q\), passes through the voxel \(q\), and experiences infinite times of reflections (see Fig. \ref{fig:mvam}). In (\ref{eq:mvam_vivid}), \(\bm\theta_q \triangleq (\mu_q, \epsilon_q)\) represents the EM parameters of voxel \(q\), and we use \( \mathcal{H}(\cdot)\) to denote the mapping from involved EM parameters \(\bm\theta\) and wavelength \(\lambda \) to the expected attenuation \(\delta\).
\subsection{Physics-Informed RF Learning}
Having established the physical model, we now describe how the underlying EM parameters  \( \bm \theta \) and radiance  \(R\)   are represented across a large-scale environment with multiple cells and grids. Let \(\bm{p}_{BS} \) be the location of the BS of a cell. Consider an EM ray from the BS to a grid along the direction \(\bm{\omega}_n\). The coordinate of the \(q\)-th voxel on this ray is
\begin{equation}
	\bm{p}_{ n,q} = \bm{p}_{BS}  + q\Delta r \bm{\omega}_n ,
\end{equation}
where \(\Delta r\) is the voxel spacing. Instead of assuming predefined materials \cite{8438326}, we learn the environment's RF parameters  directly from data. We employ coordinate-based functions \(\hat \mu(\cdot)\) and \(\hat \epsilon(\cdot)\) to map any spatial position \(\bm{p}_{n,q}\) to its estimated EM parameters, \(\bm{\hat{\theta}}_{n,q} \triangleq (\hat{\mu}(\bm{p}_{n,q}), \hat{\epsilon}(\bm{p}_{n,q}))\). The expected attenuation is then computed via our FDAM in (\ref{eq:mvam_vivid}) as
\begin{align}
	\delta_{n,q}^{\lambda} = \mathcal{H}(\bm{\hat{\theta}}_{n,q-1}, \bm{\hat{\theta}}_{n,q}, \bm{\hat{\theta}}_{n,q+1}, \lambda).
\end{align}

To model the expected radiance \(R\) of a voxel, we employ a decoupled, two-stage architecture designed to separate spatial feature extraction from view-dependent radiance synthesis.
In the first stage, a feature-encoding function, \(\mathcal{M}(\cdot)\), maps the voxel's 3D position \(\bm{p}_{n,q}\) to a high-dimensional latent feature vector, capturing its intrinsic EM properties. Subsequently, a radiance-decoding function, \(\mathcal{Q}(\cdot)\), takes this feature vector as input, conditioned on the ray direction \(\bm{\omega}_n\), grid coordinate \(\bm p_l\), and wavelength \(\lambda\), to predict the view-dependent expected radiance
\begin{equation}
	R_{n,q,l}^{\lambda} = \mathcal{Q}(\mathcal{M}(\bm{p}_{n,q}), \bm{\omega}_n, \bm{p}_l, \lambda).
\end{equation}

The architecture described above, which maps voxel coordinates to environment parameters, is conventionally realized using  MLPs. Specifically, the radiance-decoding function \(\mathcal{Q}\) is a lightweight MLP that processes the latent features and conditional inputs. The primary challenge, however, lies in the coordinate-based mapping functions \(\hat{\bm \theta}(\cdot)\), and the feature-encoder \(\mathcal{M}(\cdot)\). Implementing these with large MLPs results in significant computational overhead \cite{reiser2021kilonerf}. To overcome these limitations, we introduce an  efficient low-rank tensor-based model specifically for these three coordinate-dependent mappings in Section~\ref{sec:low-rank-section}.
\subsection{Point Cloud-Driven Regularization\label{sec:pointcloud_modeling}}
The expressive power of the coordinate-based mapping functions, combined with the ill-posed nature of the inverse problem, presents a significant challenge---the model could learn physically implausible solutions, such as creating fake scatterers in free space to fit the training data. 

To alleviate this and guide the model towards a physically meaningful representation, we introduce a strong geometric prior derived from point cloud data. The core principle is to enforce significant radiance and attenuation to happen at voxels that are located around physical obstacles.

To implement this prior, we first quantify the local geometry. We calculate the local point cloud density (PCD) of a voxel located at \(\bm p_{n,q} \),  denoted as \(D(\bm{p}_{n,q})\), by counting the number of points within the voxel spacing around the voxel's coordinate  
\begin{equation}
	D(\bm{p}_{n,q}) = \sum_{\bm{a} \in \mathcal{A}} \mathbb{I}(\|\bm{a}  - \bm{p}_{n,q}\| \leq \Delta r),
\end{equation}
where \(\mathcal{A}\) is the set of the point cloud, \(\bm a\) is the coordinate of an individual point in \(\mathcal{A}\), \(\Delta r\) is the spacing between adjacent voxels, and \(\mathbb{I}(\cdot)\) is the indicator function that returns one if its argument is true, and zero otherwise.%
Based on this density value, we then derive a modulation coefficient \(\Phi(\bm{p}_{n,q})\) using a thresholding mechanism
\begin{equation}
	\Phi(\bm{p}_{n,q})  = 
	\begin{cases} 
		1 & \text{if } D(\bm{p}_{n,q}) > D_{\text{th}} \\
		\beta \cdot D(\bm{p}_{n,q}) & \text{if } D(\bm{p}_{n,q}) \leq D_{\text{th}},
	\end{cases}
\end{equation}
where \(D_{\text{th}}\) is a density threshold and \(\beta \ll 1\) is a small constant. This coefficient, \(\Phi(\cdot)\), acts as a switch, being close to 1 in object-dense regions and close to 0 in free space.

Finally, we integrate this geometric prior directly into our model's forward pass by 
\begin{align}
	\tilde{R}_{n,q,l}^{\lambda} &= R_{n,q,l}^{\lambda} \cdot \Phi(\bm{p}_{n,q}), \label{eq:refine_R}\\
	\tilde{\delta}_{n,q}^{\lambda} &= \left(\delta_{n,q}^{\lambda}\right)^{\Phi(\bm{p}_{n,q})}. \label{eq:refine_delta}
\end{align}
The modulation in (\ref{eq:refine_R}) and (\ref{eq:refine_delta}) is physically motivated. In (\ref{eq:refine_R}), coefficient \(\Phi(\cdot)\) suppresses the radiance \(R\)  towards zero in cloud-point-free space. For attenuation \(\delta\) in (\ref{eq:refine_delta}) , the   attenuation delta approaches one in free spaces.  On the contrary, in cloud-point-dense spaces, i.e., \(\Phi=1\),  both radiance and attenuation  retain their fully learned base values.

It should be mentioned that this point cloud-driven modulation is not a post-processing step but an integral part of the model, actively pruning the solution space during training and forcing the model to attribute signal interactions to locations with physical evidence.
\subsection{ Model Optimization with Multi-domain Data}
The ultimate strength of our RF-LSCM framework lies in its ability to learn a single, shared representation of the environment's EM parameters from diverse, multi-domain data. We now formalize how this representation is optimized using data collected from a large-scale urban environment.

Consider a scenario with \(C\) cells, each operating at a specific wavelength \(\lambda_c\), serving a total of \(L\) distinct user grids. By integrating all the components developed in the preceding sections---the physically-informed FDAM (\( \delta \)), the coordinate-based mappings for EM parameters and radiance (\(R\)), and the point cloud-driven regularization---the final expected channel power for a path originating from cell \(c\) and arriving at grid \(l\) in the direction of \(\bm \omega_n\) is expressed as
\begin{equation}
	\tilde{x}_{c,l,n}^{\lambda_c}  = \left| \sum_{q=0}^Q \left( \prod_{\tilde{q}=0}^q  \tilde{\delta}_{c, n, \tilde{q}}^{\lambda_c} \right)  \tilde{R}_{c,l,n,q}^{\lambda_c}  \right|^2. \label{eq:x_final}
\end{equation}

Sampling this across all \(N\) discrete angles yields the complete APS vector 
\begin{equation}
	\bm{\tilde{x}}_{c,l}^{\lambda_c} = [\tilde{x}_{c,l,1}^{\lambda_c}, \ldots,\tilde{x}_{c,l,n}^{\lambda_c}, \ldots, \tilde{x}_{c,l,N}^{\lambda_c}]^T. \label{eq:x_vec}
\end{equation}
To learn the parameters of our model (i.e., the weights of the mapping function \(\hat{\mu}(\cdot)\), \(\hat{\epsilon}(\cdot)\), and \(\mathcal{M}(\cdot)/\mathcal{Q}(\cdot)\)), we minimize a loss function aggregated over all RSRP measurements \(\bm{r}_{c,l}^{\lambda_c}\) across cells and grids. The loss combines a data fidelity term with a sparsity-promoting regularizer
\begin{equation}
	\mathcal{L}_1 = \sum_{c=1}^C \sum_{l=1}^L\big( ||\bm{r}_{c,l}^{\lambda_c}- \mathbf{A}_{c}^{\lambda_c}\bm{\tilde{x}}_{c,l}^{\lambda_c}||_2^2 + \beta_{\text{pen}} ||\bm{\tilde{x}}_{c,l}^{\lambda_c}||_1\big). \label{eq:loss_final}
\end{equation}
The \(\ell_2\)-norm serves as a data fidelity term, while the \(\ell_1\)-norm promotes the inherent sparsity of the APS. A key advantage of our coordinate-based framework is the natural enforcement of physical consistency--- voxels shared by rays from different cells or to different grids are modeled with the same underlying properties, enabling robust multi-cell, multi-grid training.
\subsection{Penalty Design for Missing Data}\label{sec:missing_data}
While \(\mathcal{L}_1\) forms the core of our optimization, it implicitly assumes complete RSRP measurements. Unfortunately, real-world data is often imperfect. Specifically, in modern 4G/5G standards, UE conserves energy by not reporting the RSRP for beams that fall below a certain power threshold, which we denote as \(r_{\text{th}}\). This phenomenon results in a significant amount of  {systematic  missing data}. For an already challenging under-determined system (\ref{eq:yax}), this additional uncertainty about unobserved beams can severely hinder accurate APS modeling.

To counteract this, we introduce a physically-motivated penalty term that injects this prior knowledge about the reporting mechanism directly into the optimization. The penalty is designed to penalize predictions for unreceived beams only if they exceed the known reporting threshold \(r_{\text{th}}\). This is expressed as 
\begin{equation}
	\mathcal{L}_{ {P}} =\sum_{c=1}^C \sum_{l=1}^L \left\| \mathbb{I}(\bm{r}_{c,l}^{\lambda_c} = 0) \odot \left[ \mathbf{A}_{c}^{\lambda_c}\bm{\tilde{x}}_{c,l}^{\lambda_c} - r_{\text{th}} \right]_+ \right\|_2^2,
	\label{eq:penalty_loss}
\end{equation}
where \(\mathbb{I}(\cdot)\) is an element-wise indicator function. The operator \(\odot\) denotes the element-wise product. The core of the logic lies in the ReLU function, \([ \bm{x} ]_+ = \max(0, \bm{x})\). It creates a ``one-sided'' penalty: if the model's predicted RSRP (\(\mathbf{A}\bm{\tilde{x}}\)) for a missing beam is below the threshold \(r_{\text{th}}\), the penalty is zero, as this is a physically plausible reason for the data to be missing. A penalty is only incurred if the prediction violates this known constraint by exceeding the threshold.

The final, comprehensive loss function for the RF-LSCM framework is the sum of the primary objective and the missing data penalty:
\begin{equation}
	\mathcal{L}_{ } = \mathcal{L}_1 + \alpha_{P }  \mathcal{L}_{ {P }},
	\label{eq:loss_total}
\end{equation}
where \(\alpha_{P }\) is a positive penalty coefficient. By optimizing this combined objective, the model learns a representation that not only fits the observed data accurately and promotes a sparse APS, but also adheres to the physical constraints imposed by the real-world data reporting protocol.

%---
\section{{Enhancement of   RF-LSCM} \label{sec:complex_reduciton}}
As discussed in Section \ref{sec:LSCM}, estimating the channel APS from the RSRP is equivalent to solving an ill-conditioned inverse problem, making RF-LSCM difficult to be trained and perform well. In addition, in outdoor scenarios with large number of cells and grids, RF-LSCM can become computationally prohibitive due to the immense memory and processing requirements.  To address these critical challenges, this section details our approaches to reduce the computational burden of the neural field representation while enhancing model accuracy.
\subsection{Low-Rank Radiance Field Representation\label{sec:low-rank-section}}
A primary challenge in modeling radiance fields with MLPs is computational inefficiency. Querying physical parameters such as permittivity (\(\epsilon\)) and permeability (\(\mu\)), or latent features (\(\mathcal{M}\)) for every voxel requires repeated, time-consuming forward passes through large networks. To overcome this, we adopt an explicit scene representation, modeling the underlying fields as a discrete 4D tensors \cite{10.1007/978-3-031-19824-3_20} with each tensor element representing the EM parameters of a voxel in the radiance field. Thus, once the tensor is trained, the issue of querying a large MLP in the traditional radiance filed transforms to simple and efficient lookup on the tensor elements.
\subsubsection{Tensor Formulation}
We represent the radiance fields for EM parameters and latent features using two discrete 4D tensors--- \(\bm{\mathcal{G}}_\delta \in \mathbb{R}^{I \times J \times K \times D_\delta}\) and \(\bm{\mathcal{G}}_R \in \mathbb{R}^{I \times J \times K \times D_R}\). Here, \(I, J, K\) define the spatial resolution of the voxel grid along the X-, Y- and Z-axes, respectively, while \(D_\delta\) and \(D_R\) are the feature dimensions. Specifically, \(D_\delta\) is fixed at 4 to encode the real and imaginary parts of both \(\mu\) and \(\epsilon\), whereas \(D_R\) is a tunable hyperparameter for the dimension of the latent features channel.

These tensors allow for the direct retrieval of EM parameters for any voxel. For example, for a voxel located at \(\bm p_{x}\), we can obtain 
\begin{align}
	(\hat{\mu}(\bm p_{x}), \hat{\epsilon}(\bm p_{x})) &=   \bm{\mathcal{G}}_\delta(\bm p_{x}) \\
	{\mathcal{M}}(\bm p_{x}) &=  \bm{\mathcal{G}}_R( \bm p_{x})
\end{align}

\subsubsection{Efficient Structure with VM Decomposition}
We employ Vector-Matrix (VM) decomposition \cite{10.1007/978-3-031-19824-3_20}, which factorizes a tensor into a sum of outer products between vectors and matrices. For a grid of resolution \(I \times J \times K\), the components include vectors along each axis (e.g., \(\bm{v}^{X} \in \mathbb{R}^{I}\)) and matrices spanning the corresponding orthogonal planes (e.g., \(\bm{M}^{YZ} \in \mathbb{R}^{J \times K}\)), as illustrated in Fig.~\ref{fig:tensor_decompsition}.

To handle the feature channels \(D\) (e.g., \(D_\delta, D_R\)), we augment the decomposition with a small set of learnable basis vectors, \(\bm{b}_{\delta}^m \in \mathbb{R}^{D_\delta}\) and \(\bm{b}_{R}^m \in \mathbb{R}^{D_R}\) for \(m \in \{1,2,3\}\). Each of the three spatial components in the VM decomposition is then paired with one of these basis vectors, formulated as
\begin{align}
	\bm{\mathcal{G}}_{\delta} = \sum_{r=1}^{R_1} \Big( 
	& \bm{v}_{\delta, r}^{X} \circ \bm{M}_{\delta, r}^{YZ} \circ \bm{b}_{\delta}^{1} 
	+  \bm{v}_{\delta, r}^{Y} \circ \bm{M}_{\delta, r}^{XZ} \circ \bm{b}_{\delta}^{2} \nonumber \\
	+ & \bm{v}_{\delta, r}^{Z} \circ \bm{M}_{\delta, r}^{XY} \circ \bm{b}_{\delta}^{3} \Big),   \label{eq:w_corrected}
\end{align}
\begin{align}
	\bm{\mathcal{G}}_{R} = \sum_{r=1}^{R_2} \Big( 
	& \bm{v}_{R, r}^{X} \circ \bm{M}_{R, r}^{YZ} \circ \bm{b}_{R}^{1} 
	+ \bm{v}_{R, r}^{Y} \circ \bm{M}_{R, r}^{XZ} \circ \bm{b}_{R}^{2} \nonumber  \\
	+ & \bm{v}_{R, r}^{Z} \circ \bm{M}_{R, r}^{XY} \circ \bm{b}_{R}^{3} \Big), \label{eq:s_corrected}
\end{align}
where \(R_1\) and \(R_2\) are the ranks for the respective tensors, and \(\circ\) denotes the outer product. 

This low-rank structure yields a dramatic reduction in storage requirements. By choosing a low rank \(R' = \max\{R_1,R_2\}\) such that \(R' \ll \min(I, J, K)\), the parameter count is reduced from \(\mathcal O(I \cdot J \cdot K \cdot D)\) for a full 4D tensor to \(\mathcal O(R' \cdot (IJ+IK+JK))\) for the decomposed model, as the quadratic terms become dominant in large-scale scenes.  This substantial reduction enables the storage of high-resolution fields.

Beyond storage, the primary advantage of this representation is its query efficiency. A conventional mapping function implemented via the MLP is computationally expensive; a forward pass through a network with \(M'\) layers of width \(W\) has a computational complexity of \(\mathcal O(M' \cdot W^2)\) for every voxel queried. In   contrast, our tensor-based query—which primarily involves interpolation and summing the \(R'\) low-rank components—has a much lower complexity of \(\mathcal O(R' \cdot D)\). Collectively, this dual advantage in storage and speed empowers our model to represent large-scale, high-resolution fields in a way that is both memory-conscious and computationally efficient.

\subsection{Hierarchical Tensor Angular Modeling (HiTAM)}

It is important to point out that the angular resolution of channel APS (i.e., \(N\)) is a key factor that causes high training and inference complexity. In particular, the rendering of all voxels from \(N\) angles in (\ref{eq:x_final}) lead to extremely high computation burden.  Unfortunately, in practice \(N\) is  typically large.  
For instance, optimizing antenna configurations often requires discretizing tilt and azimuth into high-grained resolutions, with 91 and 72 discrete levels for the tilt and azimuth axes respectively, resulting in \(N=6552\) angular dimensions. Modeling such a high-dimensional space, especially in large-scale outdoor scenes, far exceeds the memory and compute capacity of typical hardware. To overcome this bottleneck, we propose the HiTAM framework, a novel coarse-to-fine strategy that integrates our low-rank RF model with adaptive angular sampling.

As illustrated in Fig.~\ref{fig:c2f}, HiTAM includes two RF-LSCM models, one with a coarse angular resolution and the another one with a refined angular resolution. It consists of two hierarchical stages. 
\begin{figure}[t]
	\centering
	\includegraphics[width=1 \linewidth]{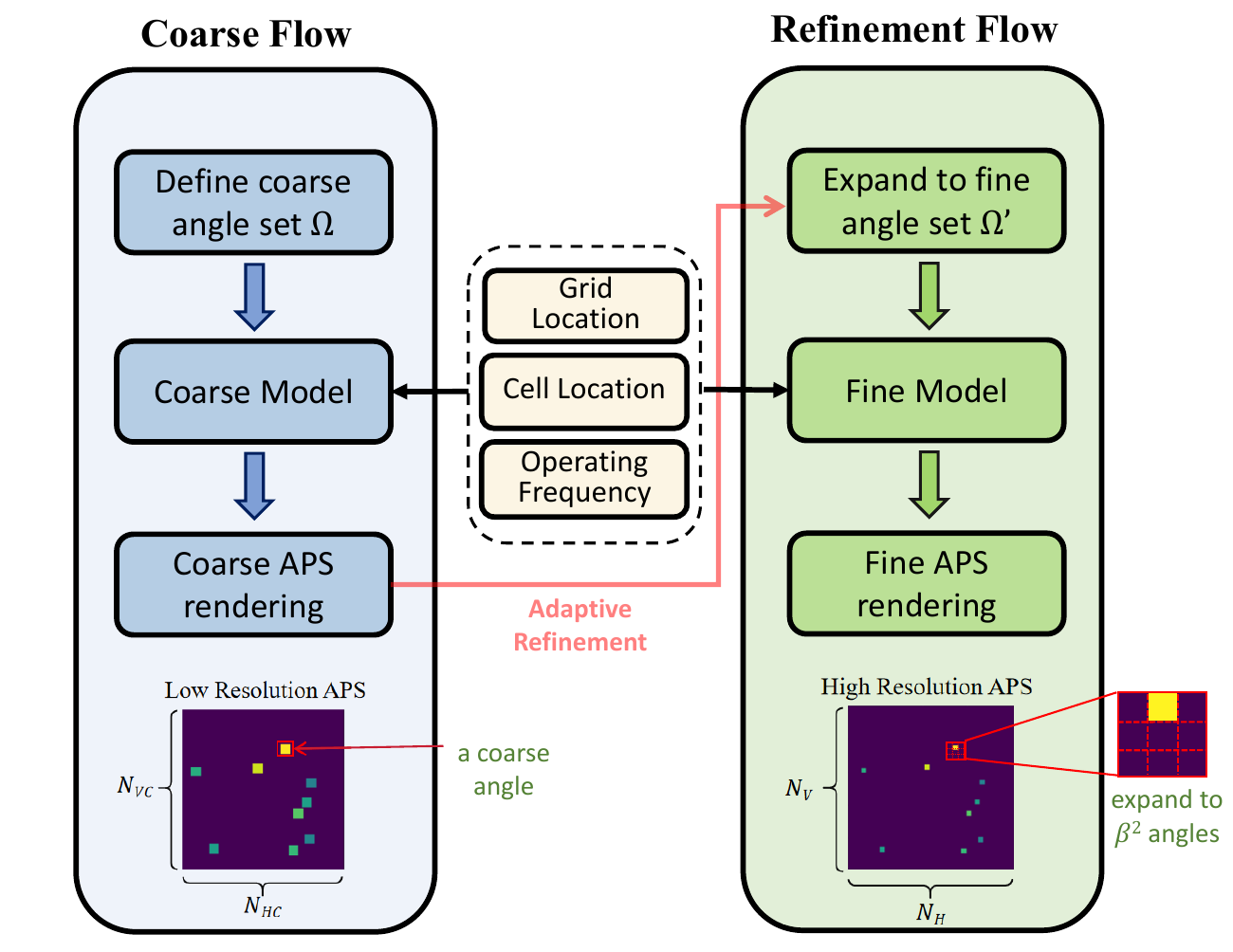} %c2f_new
	\caption{The Flow chart of HiTAM. Both Coarse and Refinement models are low-rank RF tensor models. \(K_C\)   coarse angles are discretized into \(\beta^2 K_C\) fine angles for the Refinement Flow.}
	\label{fig:c2f}
	%	\Description{A visual representation of the Hierarchical Tensor Angular Modeling (HiTAM) framework for wireless signal propagation modeling.}
\end{figure}

\textbf{Stage 1: Coarse Flow.}
We define a coarse angle set \( {\Omega} \) with a resolution of \(N_{VC} = \beta^{-1}N_V\) and \(N_{HC} = \beta^{-1}N_H\), where \(\beta > 1\) is an integer downsampling factor, and  \( |{\Omega}| = N_{VC} \times N_{HC}\). A Coarse model then renders the coarse APS, denoted as \(\bm{\tilde{x}}^{(1)} \in \mathbb{R}^{N^{(1)}}\) (where the indices \(c,p,\lambda_c\) are omitted for notational simplicity).

\textbf{Stage 2: Adaptive Refinement  Flow.}
Instead of wastefully refining all angles, this second stage focuses computational resources exclusively on the candidate directions identified in Stage 1. We first select the top \(K_C\) angles from the coarse APS \(\bm{\tilde{x}}^{(1)}\) based on their power strength. Each of these \(K_C\)  angle is then expanded into a finer \(\beta \times \beta\) sub-grids. The union of these sub-grids forms the final, sparse high-resolution set of angles, \( {\Omega}'\). A separate Refinement model then processes only the rays corresponding to this reduced set (where \(| {\Omega}' | = \beta^2 K_C \ll N_V N_H\)) to compute the high-resolution, sparse APS, denoted as \(\bm{\tilde{x}}^{(2)}\). This adaptive approach allows HiTAM to achieve high-resolution modeling with tightly controlled computational complexity.

\textbf{Joint Optimization.}
The two stages are trained jointly using a composite loss function that evolves during training. For simplicity, cell and grid indices are omitted. The total loss is a weighted sum of the losses from both stages
\begin{equation}
	\begin{aligned}
		\mathcal{L}_{\text{total}} &= (1-\alpha)\mathcal{L}^{(1)} + \alpha\mathcal{L}^{(2)} 
		\ \\
		\text{where} \quad \mathcal{L}^{(1)} &= \|\bm{r}-\mathbf{A}^{(1)}\bm{\tilde {x}}^{(1)}\|_2^2 + \beta_\text{pen}\|\bm{\tilde {x}}^{(1)}\|_1 +   \alpha_{P } \mathcal{L}_{ {P }}^{(1)}, \\
		\mathcal{L}^{(2)} &= \|\bm{r}-\mathbf{A}^{(2)}\bm{\tilde {x}}^{(2)}\|_2^2 + \beta_\text{pen}\|\bm{\tilde {x}}^{(2)}\|_1
		+   \alpha_{P }  \mathcal{L}_{ {P }}^{(2)},
	\end{aligned}
\end{equation}
where \(\alpha\) is a curriculum learning parameter that progressively increases during training, gradually shifting focus from the coarse estimation to the  refinement. The matrices \(\mathbf{A}^{(1)}\) and \(\mathbf{A}^{(2)}\) are sub-matrices of the full system matrix \(\mathbf{A}\), constructed by selecting columns corresponding to the angles in \( {\Omega} \) and \( {\Omega}'\), respectively.  
The terms \(\mathcal{L}_P^{(1)}\) and \(\mathcal{L}_P^{(2)}\) represent the penalty losses defined in (\ref{eq:penalty_loss}), computed on \(\bm{\tilde {x}}^{(1)}\) and \(\bm{\tilde {x}}^{(2)}\), respectively. 

\section{Evaluation \label{sec:evaluation}}

 \subsection{{RF-LSCM Implementation  }\label{sec:impletmentation}}
 
 The RF-LSCM framework is implemented using PyTorch. The setting of batch size is 256 to accommodate the high granularity required by APS. The Coarse Model is initially trained for 1000 iterations before being integrated with the Refinement Model for joint training. We utilize the Adam optimizer \cite{kingma2017adammethodstochasticoptimization} with an initial learning rate of 0.01 for tensor factors and 0.001 for the MLP-based network \(\mathcal R \), while retaining all other hyperparameters at their default settings. The model is trained on a single NVIDIA 4090 GPU.
	
\subsection{Experiment on Collaborative Multi-Cell Channel Modeling\label{sec:COLLABORATIVE}}

% 请确保在文档开头加载了 \usepackage{graphicx}, \usepackage{caption}, \usepackage{subcaption}

% 在您的文档导言区 (preamble)，请确保加载了以下宏包
% \usepackage{graphicx}
% \usepackage{caption}
% \usepackage{subcaption}
% \usepackage{adjustbox} % 建议添加此宏包，它提供了更强的对齐工具

In this study, we contemplate employing real world data from multiple cells to model the  channel across multi-cell.
% 确保你的导言区已经加载了 \usepackage[caption=false,font=footnotesize]{subfig}
% 并且没有加载 subcaption 和 caption 宏包

\begin{figure*}[!t] % [!t] 是IEEE模板中推荐的跨双栏图表放置选项，表示强制顶部放置
    \centering
    % 第一个子图
    \subfloat[Multi-Cell 3D Map.]{\includegraphics[width=0.45\linewidth]{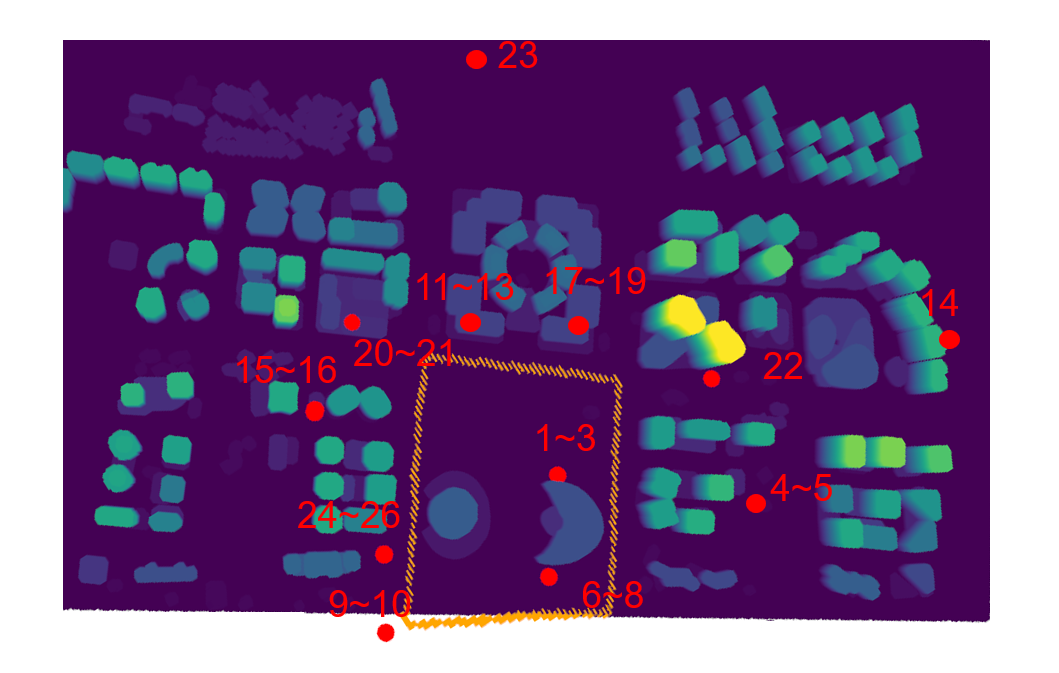}%
    \label{fig:hangzhoumap}}
    \hfill % 这个命令会在两个子图之间产生弹性空白，使它们左右对齐
    % 第二个子图
    \subfloat[Prediction Performance on Multi-cell RSRP.]{\includegraphics[width=0.45\linewidth]{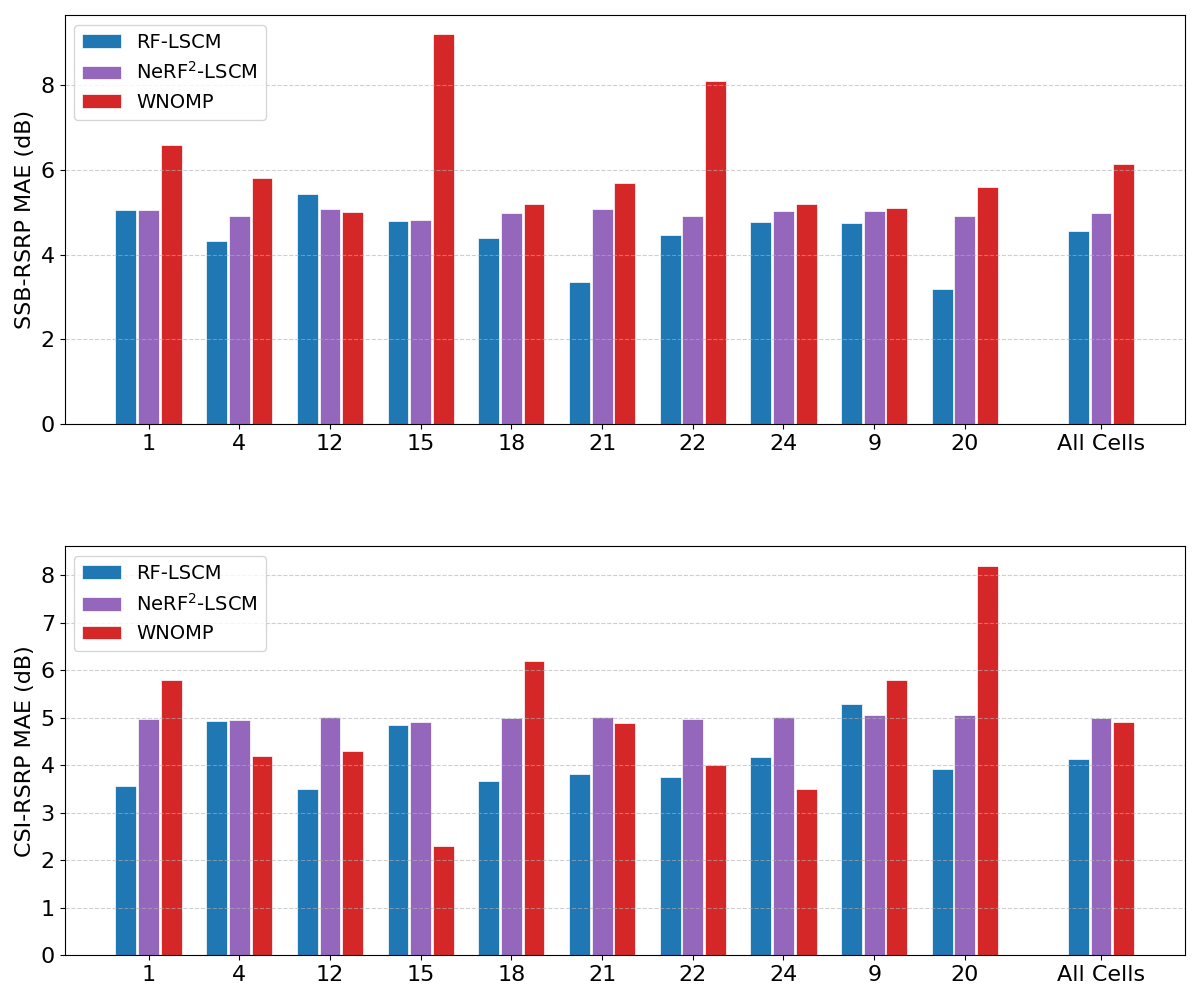}%
    \label{fig:hangzhouresult}}
    
    \caption{\textbf{Evaluation of RF-LSCM on Real-World Multi-Cell Data}. RF-LSCM effectively conducts multi-cell collaborative modeling using drive test data, outperforming the baseline method with reduced prediction errors after antenna adjustment.}
    \label{fig:Multi-Cell Evaluation}
\end{figure*}
%\subsubsection{Multi-Cell Modeling Experiment Setup}
%In Figure \ref{fig:Multi-Cell Evaluation}, we present a real world three-dimensional map of an urban environment consisting of 26 cells, each denoted by red circles. 
%The antenna of the BS  is AAU5619 \cite{AAU5619} at 2.6GHz. We collect real world RSRP data using a smartphone equipped with the PHU APP \cite{huawei_probe_handset_unit}. This application enables us to gather data over an extended period, thereby allowing us to calculate the average RSRP over a sufficient time slot.
%The measurements were taken around the buildings within the yellow box, utilizing a dataset that includes both drive test data collected from real-world scenarios. By adjusting specific antennas within these cells, we are able to collect 2 rounds the data before and after the adjustment. We then trained the model using the on round of the dataset, and test in the another round data. 

\subsubsection{Multi-Cell Modeling Experiment Setup}
In Fig.~\ref{fig:Multi-Cell Evaluation}\subref{fig:hangzhoumap}, we present a real-world three-dimensional map of an urban environment consisting of 26 cells, and the BS of each cell is marked by red points. 
Drive tests were conducted within the area outlined by the yellow box in the figure to collect real-world multi-beam RSRP data. 
The dataset comprises approximately 800 grids (each measuring 10 m × 10 m). 

As we were unable to obtain groundtruth APS measurements, the model's evaluation was performed in two stages across \(L\) distinct grid points. First, we collected an initial set of RSRP measurements, \(\{\bm{r}_l\}_{l=1}^L\), from these \(L\) grids. Our method then used this set to construct the corresponding APS for each grid, \(\{\bm{\tilde{x}}_l\}_{l=1}^L\). Subsequently, the antennas were adjusted to generate a new sensing matrix \(\mathbf{A'}\), and a second set of RSRP measurements, \(\{\bm{r'}_l\}_{l=1}^L\), was collected from the same \(L\) locations. The MAE between the RSRPs predicted for the second round and the actual measurements was then calculated as
\begin{equation}
\text{MAE} = \frac{1}{L} \sum_{l=1}^{L} \|10\log_{10}\bm{r'}_l -10\log_{10} \mathbf{A'}\bm{\tilde{x}}_l\|_1 .
\end{equation}
%
%By adjusting specific antennas within these cells, we were able to collect data in two rounds--- before and after the adjustment. We then trained the model using one round of the dataset and tested it on the other round of data.
%
In this context, we introduce the baselines of our experiment.
\begin{itemize}
	\item WNOMP (Weighted Non-negative Orthogonal Matching Pursuit \cite{10299600}): WNOMP is an improved OMP-type algorithm that aims to solve the sparse recovery problem (\ref{eq:muticell}).
	\item NeRF$^2$-LSCM: We adopt the MLP architecture as in NeRF$^2$ \cite{Zhao_2023} to build the APS in our LSCM flow.
 \end{itemize}

\subsubsection{Evaluation of Collaborative Multi-cell Modeling}

We adjust the antenna parameters using two distinct methods, each impacting the measurement matrices and received signal strengths  differently
\begin{itemize}
	\item \textbf{Method (a): Modifying the Antenna Codebook.} This approach fundamentally alters the structure of the sensing matrix \(\bf A\), causing significant variations in the SSB-RSRP. It is applied to cells 1, 4, 12, 15, 18, 21, 22, and 24.
	
	\item \textbf{Method (b): Physical Antenna Rotation.} This method involves adjusting the antenna's angle, which induces cyclic shifts in the rows and columns of matrix \(\bf A\). This leads to substantial variations in both SSB-RSRP and CSI-RSRP and is used for cells 9 and 20.
\end{itemize}

As shown in Fig.~\ref{fig:Multi-Cell Evaluation}\subref{fig:hangzhouresult}, the MAE of the prediction errors for RF-LSCM is significantly lower than that of the baseline method (WNOMP) in both SSB-RSRP and CSI-RSRP scenarios. For SSB-RSRP, our method achieves an MAE of 4.27 dB, compared to 6.10 dB for the baseline method and 4.97 dB for NeRF$^2$-LSCM. This represents a \textbf{1.83} dB improvement over the baseline method, equivalent to a \textbf{30\%} performance enhancement across all cells.

Regarding CSI-RSRP, the MAE across all cells was 4.14 dB for our method, 4.92 dB for the baseline method, and 4.99 dB for NeRF$^2$-LSCM. Since CSI-RSRP is predominantly affected by adjustment method (b), we specifically examined the MAE for cells 9 and 20. Our method achieved an MAE of 5.25 dB for these two cells, compared to the baseline method's 7.01 dB, reflecting a \textbf{1.75} dB improvement.

In addition to its superior accuracy, RF-LSCM exhibits a dramatic advantage in training efficiency. Our model converges in approximately 20 minutes, making it an order of magnitude faster than NeRF²-LSCM, which requires around 200 minutes.}

The subpar performance of NeRF²-LSCM stems from its inherent difficulty in solving the ill-posed inverse problem of recovering a high-angular-resolution APS from low-dimensional RSRP measurements. This fundamental limitation leads to significant training instability, which is evidenced by the large standard deviation observed across the learning curves from multiple independent runs (see ablation experiment in Section~\ref{sec:ablation_hitam}). In contrast, our method overcomes this instability by employing a two-stage HiTAM algorithm. By progressively constructing the APS in a coarse-to-fine manner, our approach achieves a robust solution to the ill-posed problem, yielding precise channel models that are better adapted to complex urban environments.

% \subsubsection{\textcolor{red}{RSRP Prediction For Unexplored Location}}
% Traditional methods such as WNOMP have demonstrated limitations in predicting the APS when the RSRP of such grid is unknown. Our proposed model harnesses the capabilities of neural networks to predict APS without relying on prior RSRP knowledge within certain grid areas. This advancement implies that our method is capable of modeling channel APS for grid areas where RSRP is currently unknown, thereby facilitating the assessment of communication quality.

% To validate this, we designed an experiment where the test data, collected after an antenna adjustment, was composed of two distinct types of locations. The first type included grids that were also present in the training dataset, while the second, more critical type, consisted of grids whose coordinates were entirely excluded from the training dataset. Our analysis focused on this second group of 141 previously unseen grids across 10 cells. Utilizing our model, we predict the APS for these new grids and subsequently predicted the RSRP after antenna adjustment. The results show that we achieved a MAE of 5.15 dB, demonstrating that our model has achieved a high level of accuracy in predicting communication coverage quality within the network for unexplored locations.

% 再次确认您的导言区已加载 \usepackage[caption=false,font=footnotesize]{subfig}
% 并且没有加载 subcaption 和 caption 宏包

\begin{figure*}[!t] % 使用 [!t] 强制图片置于页面顶部
	\centering
	
	% 第一个子图 (a)
	\subfloat[Data distribution of the collected multi-frequency dataset.]{\includegraphics[width=0.45\linewidth, height=0.35\textheight, keepaspectratio]{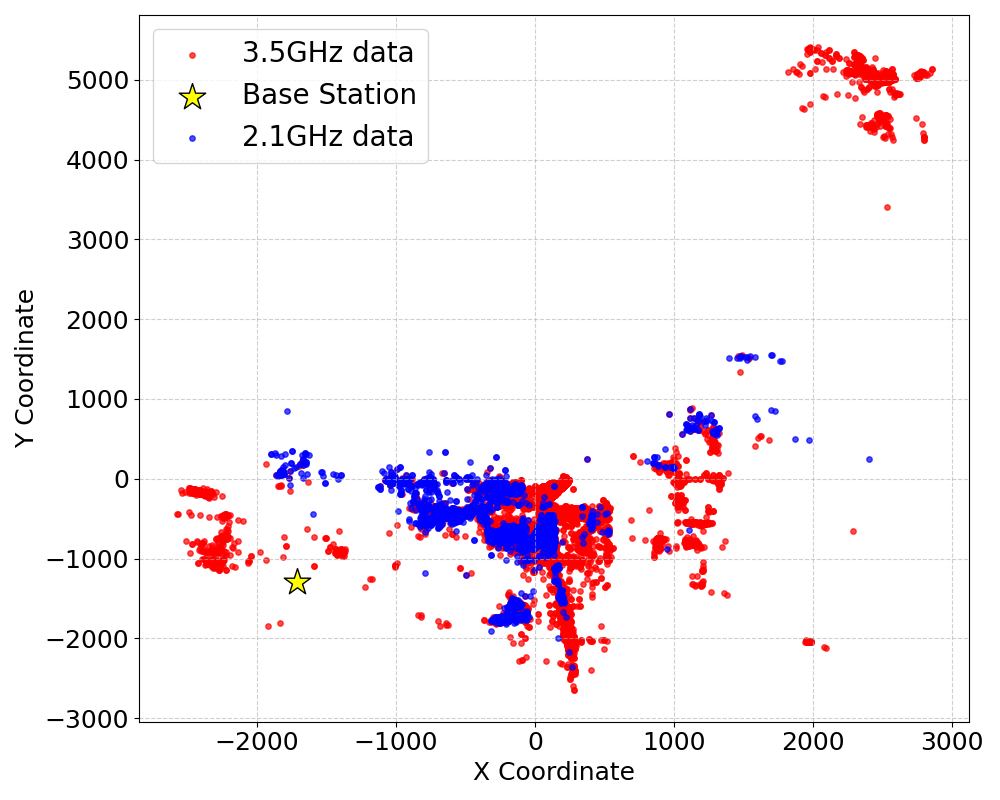}%
	\label{fig:multifreqdata}}
	\hfill % 在子图之间添加水平弹性空白
	% 第二个子图 (b)
	\subfloat[Performance comparison as a function of the percentage (p\%) of 2.1 GHz data used in training.]{\includegraphics[width=0.45\linewidth, height=0.35\textheight, keepaspectratio]{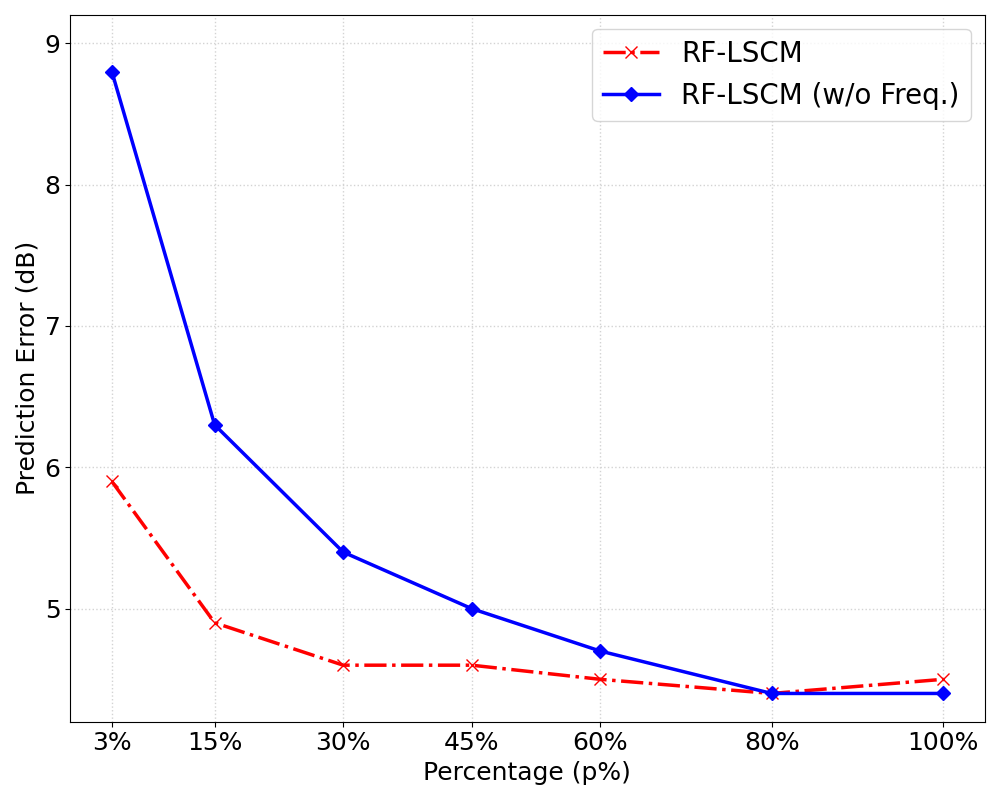}%
	\label{fig:multifreqexp}}
	
	\caption{Analysis of multi-frequency channel modeling. (a) The distribution of the collected 2.1 GHz and 3.5 GHz datasets. (b) The impact of augmenting the training set with varying amounts of 2.1 GHz data (with/without 3.5GHz data).}
	\label{fig:multi_freq_combined}
\end{figure*}

\subsection{Multi-Frequency Channel Modeling}

The core objective of this experiment is to demonstrate that our RF-LSCM architecture can leverage multi-frequency information to enhance prediction performance. To achieve this, we collected data from six nearby base stations operating at 2.1 GHz and 3.5 GHz. As illustrated in Fig.~\ref{fig:multi_freq_combined}\subref{fig:multifreqdata}, the spatial distributions for the two frequencies are not perfectly aligned, reflecting a realistic data collection scenario.

To isolate and quantify the benefit of using out-of-band data, we designed an experiment with a specific training and testing setup. First, we designated 70\% of the 2.1 GHz data as a fixed test set. The remaining 30\% of the 2.1 GHz data formed a training pool. We then compared two models:

\begin{enumerate}
	\item \textbf{RF-LSCM (w/o Freq.):} A baseline model trained using only a limited fraction (p\%) of the 2.1 GHz training pool. This scenario measures performance using only scarce, in-band information.
	\item \textbf{RF-LSCM:} Our full model, trained on the same scarce p\% of 2.1 GHz data, but augmented with the entire 3.5 GHz dataset.
\end{enumerate}

By evaluating both models on the 2.1 GHz test set, we can directly measure the performance gain attributable to the model's ability to use the 3.5 GHz data. The results, presented in Fig.~\ref{fig:multi_freq_combined}\subref{fig:multifreqexp}, strongly support our hypothesis. The full RF-LSCM model consistently achieves a lower MAE across all values of p. The advantage is most pronounced when 2.1 GHz data is scarcest; for instance, at p=3\%, our model improves the MAE by 21.8\% (from 8.7dB to 6.8dB). This result not only shows a significant performance improvement but also demonstrates that our model's architecture successfully extracts and transfers knowledge across different frequency bands.
\section{{Ablation Study}} \label{sec:ablation_study}
In this section, we conduct a series of ablation studies to demonstrate the effectiveness of the key components within our proposed model. The dataset used in this section is the same as the Section~\ref{sec:evaluation}.
% 
% 请确保导言区已加载 \usepackage[caption=false,font=footnotesize]{subfig}
% 并且没有加载 subcaption 和 caption 宏包

% 请确保导言区已加载 \usepackage[caption=false,font=footnotesize]{subfig}
% 并且没有加载 subcaption 和 caption 宏包
% =======================================================
%  FIGURE 1: SINGLE-COLUMN FIGURE (Original Subfigure a)
% =======================================================

% \begin{figure}[!t] % 使用 [!t] 建议 LaTeX 将其放置在栏的顶部
%     \centering
%     % 对于单栏图，宽度设为 0.9\linewidth 是个不错的选择
%     \includegraphics[width=0.9\linewidth]{plot/MAE_missingRate}
%     \caption{RSRP imputation MAE across varying missing rates, demonstrating the model's robustness to data loss.}
%     \label{fig:myplot} % 标签保持不变
% \end{figure}
% % =======================================================
% %  FIGURE 2: DOUBLE-COLUMN FIGURE* (Original Subfigures b and c)
% % =======================================================

\begin{figure*}[!t] % 跨双栏图，强制顶部放置
    \centering
    
    % 新的子图 (a)，内容来自原来的子图 (b)
    \subfloat[Performance curves highlight the importance of HiTAM for convergence and lower MAE.]{\includegraphics[width=0.48\textwidth]{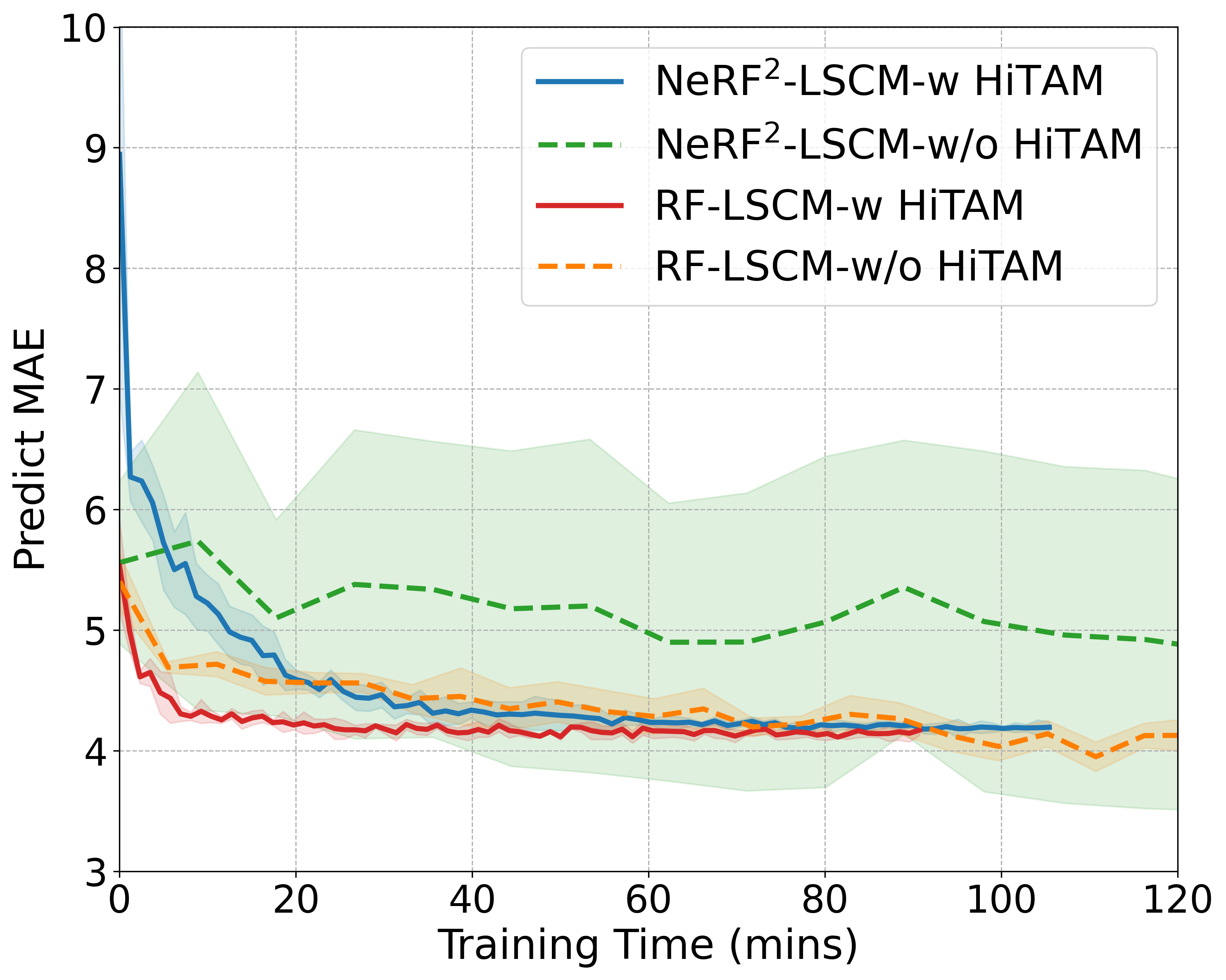}%
    \label{fig:nerf-rflscm-wohitam}} % 内部标签保持不变
    \hfill % 水平分隔
    % 新的子图 (b)，内容来自原来的子图 (c)
    \subfloat[Impact of parameter $\alpha_P$ on RF-LSCM performance, showing the efficiency of the penalty design.]{\includegraphics[width=0.48\textwidth]{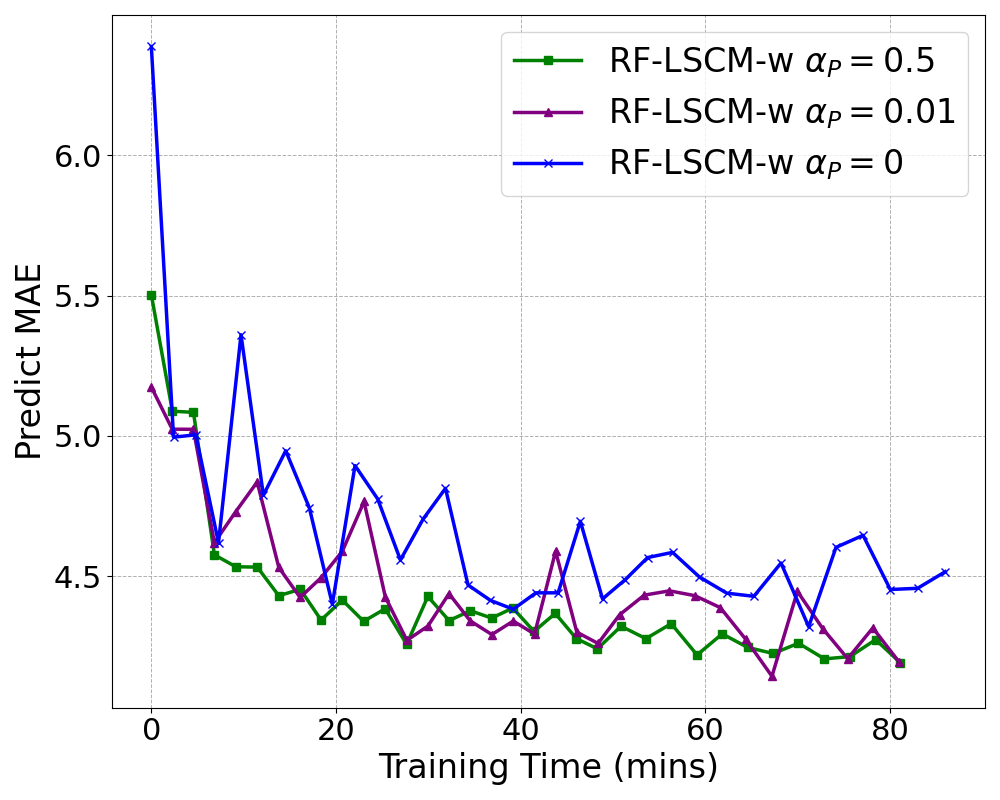}%
    \label{fig:ablation_alpha_p}} % 内部标签保持不变
    
    % 为这个新的组合图创建一个新的、更合适的总标题
    \caption{Ablation studies for RF-LSCM. (a) Performance and convergence comparison of model variants, highlighting the role of HiTAM. (b) Performance variation with different values of the penalty parameter $\alpha_P$.}
    \label{fig:ablation_studies_combined} % 创建一个新的总标签
\end{figure*}
% \begin{figure*}[!t] % 使用 [!t] 强制图片置于页面顶部
% 	\centering
	
% 	% 第一个子图 (a)
% 	\subfloat[RSRP imputation MAE across varying missing rates, demonstrating the model's robustness to data loss.]{\includegraphics[width=0.32\textwidth]{plot/MAE_missingRate}%
% 	\label{fig:myplot}}
% 	\hfill % 在第一和第二个子图之间添加水平弹性空白
% 	% 第二个子图 (b)
% 	\subfloat[Performance curves highlight the importance of HiTAM for convergence and lower MAE.]{\includegraphics[width=0.32\textwidth]{plot/nerf-rflscm-wo_HiTAM}%
% 	\label{fig:nerf-rflscm-wohitam}}
% 	\hfill % 在第二和第三个子图之间添加水平弹性空白
% 	% 第三个子图 (c)
% 	\subfloat[Impact of parameter $\alpha_P$ on RF-LSCM performance, showing the efficiency of the penalty design.]{\includegraphics[width=0.32\textwidth]{plot/L_p plot4}%
% 	\label{fig:ablation_alpha_p}}
	
% 	\caption{Comprehensive analysis of RF-LSCM: (a) RSRP Imputation MAE under different missing rates, (b) Performance and convergence comparison of model variants, and (c) Performance variation with different values of $\alpha_P$.}
% 	\label{fig:combined}
% \end{figure*}
\subsection{Ablation Study of Tensor-based RF Channel Representation and HiTAM} \label{sec:ablation_hitam}
To validate the effectiveness of the two core components in our proposed method—1) the tensor-based RF channel representation and 2) the two-stage HiTAM module—we conducted a comprehensive ablation study. We compared four distinct model variants, evaluating their performance using MAE. The models under comparison include:
\begin{itemize}
	\item \textbf{RF-LSCM (-w HiTAM):} Our full proposed model.
	\item \textbf{RF-LSCM-w/o HiTAM:} Our model with the HiTAM module removed, isolating the performance of our tensor representation alone.
	\item \textbf{NeRF²-LSCM-w  HiTAM:} The baseline NeRF² representation combined with our HiTAM module.
	\item \textbf{NeRF²-LSCM-w/o HiTAM:} The baseline NeRF² representation without HiTAM, serving as a foundational baseline.
\end{itemize}

For this experiment, all models were trained on data from one round and evaluated on data from a different, unseen round. The MAE was recorded every 10 epochs to track convergence. 
The convergence results are presented in Fig.~\ref{fig:ablation_studies_combined}\subref{fig:nerf-rflscm-wohitam}, where lines depict the MAE over training time, and the shaded regions represent the standard deviation across multiple runs. The results unequivocally demonstrate the superiority of our full model, RF-LSCM.

Specifically, RF-LSCM achieves a low MAE of 4.2 dB in just 20 minutes, exhibiting minimal variance that underscores its high training stability. This performance advantage stems from how our model tackles the challenging recovery problem formulated in (\ref{eq:muticell}). Directly reconstructing a high-dimensional APS is an inherently difficult, ill-posed problem that leads to training instability, as evidenced by the models lacking our HiTAM module. The pure MLP-based architecture (NeRF²-LSCM-w/o HiTAM) fails to converge effectively, displaying a high MAE and an extremely broad standard deviation. Concurrently, while the tensor-based architecture without HiTAM (RF-LSCM-w/o HiTAM) eventually reaches a low MAE, its convergence time of 110 minutes is prohibitively long. Our HiTAM module directly mitigates this issue by employing a two-stage, coarse-to-fine estimation process. This approach effectively reduces the dimensionality of the problem at each stage, ensuring both rapid and robust convergence.

Furthermore, isolating the contribution of the tensor-based representation is achieved by comparing RF-LSCM with NeRF²-LSCM-w HiTAM. Although the baseline with HiTAM reaches a 4.2 dB MAE, it requires 75 minutes. The fact that our RF-LSCM is nearly four times faster validates that our proposed tensor representation for RF channels is inherently more efficient and effective for this task than the baseline NeRF$^2$ approach.
 
\subsection{Ablation Study of Penalty coefficient design for Missing Data}
This section evaluates the effectiveness of the penalty term, defined in  (\ref{eq:penalty_loss}), for modeling the channel's APS. To investigate the effect of different penalty coefficients, \(\alpha_{P}\), on model performance, we present an ablation study in Fig.~\ref{fig:ablation_studies_combined}\subref{fig:ablation_alpha_p}. The results indicate that setting \(\alpha_{P} = 0.5\) allows the model to achieve robust and consistent performance across the training epochs.

\subsection{Ablation Study of Point Clouds Regularization}
\begin{figure}[t]
	\centering
	\includegraphics[width=1\linewidth]{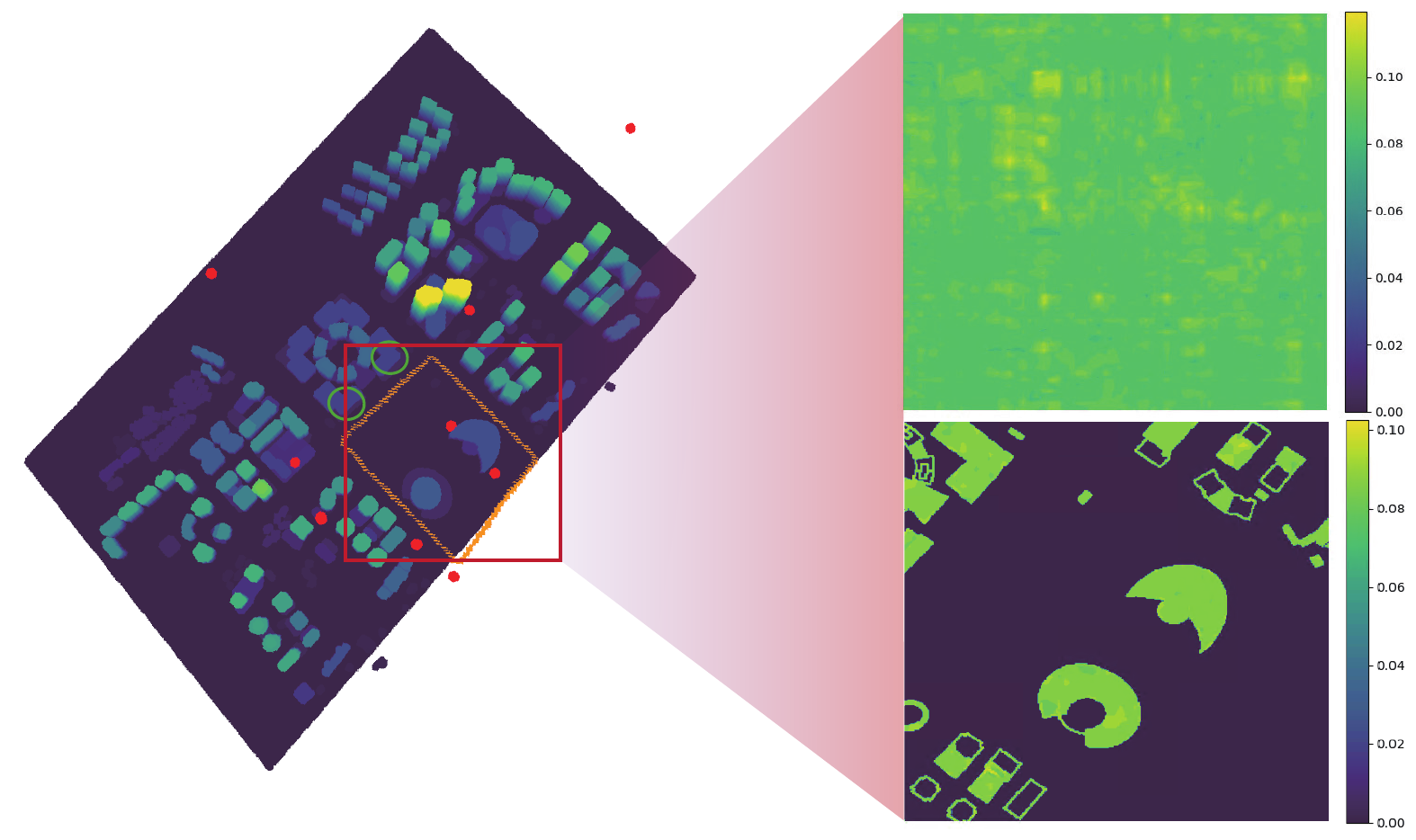 }
	\caption{\textbf{Expected Attenuation Map}. The attenuation map predicted by RF-LSCM with (bottom)/without (top) point clouds regularization. The shown attenuation \(\overline {\delta}  = -\log(\delta)\)   exhibits high consistency with the buildings, demonstrating the model's consistency with physical reality.}
	\label{fig:deltahanghzou}
\end{figure}

To evaluate the model's efficacy of point cloud regularization, we analyze the predicted attenuation map. As shown in Figure \ref{fig:deltahanghzou}, which presents a top-down view of the primary server area, we visualize the expected attenuation coefficient \(\delta\). For enhanced clarity, the map is plotted using \(\overline{\delta} = -\log(\delta)\), where higher values indicate stronger attenuation. The results show a strong correspondence between the predicted attenuation and the actual building layouts, confirming the model's consistency with physical reality. This physical consistency is significantly improved by our proposed point cloud regularization, as evidenced by the comparison between the model with it (bottom) and without it (top). 
The value of this regularization is twofold. Beyond the visual improvement, it yields a significant performance boost, lowering the MAE from 5.0 dB to 4.5 dB when trained with just 10\% of the training set. This result demonstrates that leveraging point cloud data can effectively guide channel modeling, particularly in low-data scenarios.

\section{CONCLUDING REMARKS\label{sec:concluding}}
We have introduced RF-LSCM, a novel framework that advances localized statistical channel modeling by addressing the limitations of traditional methods in handling multi-cell, multi-grid, and multi-frequency scenarios. RF-LSCM leverages a radiance field representation to directly model the channel APS, incorporating a proposed  FDAM  to accurately capture signal propagation across diverse spatial and spectral conditions.
To ensure computational tractability, RF-LSCM employs a vector-matrix low-rank tensor decomposition  coupled with a HiTAM algorithm. This approach significantly reduces computational overhead while maintaining high fidelity in channel representation.
Experimental validation on  real-world datasets demonstrated RF-LSCM's superiority over state-of-the-art methods. Notably, it achieved up to a 30\% reduction in MAE for coverage prediction and a 21.8\% MAE reduction through multi-frequency data fusion. These findings highlight RF-LSCM's efficacy and potential as a robust tool for precise cellular network optimization and the design of future  wireless networks.

	%\begin{acks}
	%	Need revises
	%\end{acks}
	
	
	\bibliographystyle{IEEEtran}
	%%\bibliography{sample-base}
	\bibliography{references}
	
	%\appendix
	
	%Lorem ipsum dolor sit amet, consectetur adipiscing elit. Morbi
	%malesuada, quam in pulvinar varius, metus nunc fermentum urna, id
	%sollicitudin purus odio sit amet enim. Aliquam ullamcorper eu ipsum
	%vel mollis. Curabitur quis dictum nisl. Phasellus vel semper risus, et
	%lacinia dolor. Integer ultricies commodo sem nec semper.
	%Etiam commodo feugiat nisl pulvinar pellentesque. Etiam auctor sodales
	%ligula, non varius nibh pulvinar semper. Suspendisse nec lectus non
	%ipsum convallis congue hendrerit vitae sapien. Donec at laoreet
	%eros. Vivamus non purus placerat, scelerisque diam eu, cursus
	%ante. Etiam aliquam tortor auctor efficitur mattis.
	%Nam id fermentum dui. Suspendisse sagittis tortor a nulla mollis, in
	%pulvinar ex pretium. Sed interdum orci quis metus euismod, et sagittis
	%enim maximus. Vestibulum gravida massa ut felis suscipit
	%congue. Quisque mattis elit a risus ultrices commodo venenatis eget
	%dui. Etiam sagittis eleifend elementum.
	%
	%Nam interdum magna at lectus dignissim, ac dignissim lorem
	%rhoncus. Maecenas eu arcu ac neque placerat aliquam. Nunc pulvinar
	%massa et mattis lacinia.

\end{document}